\documentclass{article}

\newcommand{\OverallComparisonTableSmall}{{
\begin{table}[t]
\centering
\captionof{table}{
Performance ($\uparrow$) comparison of QAT, QaRL, QUADS, and \sysname under joint NVFP4 W/A/KV rollout setting on Qwen3.5-35B-A3B for reasoning RL tasks. The best performance is marked in bold. The performance gain compared to baselines is marked in green inside bracket.
}
\vspace{-2mm}

\resizebox{1\textwidth}{!}{%
\begin{tabular}{l|l|cccc|c}
\midrule
&
& \multicolumn{5}{c}{\textsc{Qwen3.5-35B-A3B}}
\\
\cmidrule(lr){3-7}

\textsc{Rollout Config}
& \textsc{Algo.}
& \textsc{LiveCodeBench}
& \textsc{AIME24}
& \textsc{AIME25}
& \textsc{HMMT25}
& \textsc{Average}
\\
\midrule

\textcolor{mygray}{BF16 W/A + BF16 KV}
& \textcolor{mygray}{Vanilla}
& \textcolor{mygray}{\textbf{67.1}}
& \textcolor{mygray}{83.8}
& \textcolor{mygray}{\textbf{81.3}}
& \textcolor{mygray}{67.5}
& \textcolor{mygray}{74.9}
\\\midrule

\multirow{4}{*}{NVFP4 W/A + NVFP4 KV}
& QAT
& 55.1
& 70.7
& 63.4
& 49.2
& 59.6
\\
\cmidrule{2-7}

& QaRL
& 54.1
& 73.3
& 57.9
& 49.2
& 58.6
\\
\cmidrule{2-7}

& QUADS
& 60.4
& 80.5
& 75.4
& 59.0
& 68.8
\\
\cmidrule{2-7}

& \textbf{\sysname}
& 66.4 (\textcolor{mygreen}{$\uparrow$6.0}) 
& \textbf{86.3} (\textcolor{mygreen}{$\uparrow$5.8}) 
& 78.5 (\textcolor{mygreen}{$\uparrow$3.1}) 
& \textbf{70.0} (\textcolor{mygreen}{$\uparrow$11.0}) 
& \textbf{75.3} (\textcolor{mygreen}{$\uparrow$6.5}) 
\\

\midrule
\end{tabular}
}
\vspace{-4mm}
\label{tab:overall_comparison_35b}
\end{table}
}}

\newcommand{\OverallComparisonTableLarge}{{
\centering
\captionof{table}{
Performance ($\uparrow$) on larger-scale MoE LLMs, including Qwen3.5-122B-A10B and Qwen3.8-Flash-Next (125B-A6B) trained on coding RL tasks and evaluated on DeepSWE and Terminal-Bench, respectively. Qwen3.8-2.4T-A95B is trained on long-horizon RL task and evaluated on GDPval. The best score in each row is marked in bold. The performance gain compared to baselines is marked in green inside bracket.
}
\vspace{-2mm}

\resizebox{0.55\textwidth}{!}{%
\begin{tabular}{l|c|cc|c}
\midrule

\textsc{Model}
& \textcolor{mygray}{\textsc{BF16}}
& \textsc{QAT}
& \textsc{QUADS}
& \textsc{\sysname}
\\
\midrule

\textsc{Qwen3.5-122B-A10B}
& \textcolor{mygray}{\textbf{33.4}}
& 28.8 & 29.1
& 33.0 (\textcolor{mygreen}{$\uparrow$3.9})
\\

\textsc{Qwen3.8-Flash-Next}
& \textcolor{mygray}{68.8}
& 60.4 & 66.4
& \textbf{70.6} (\textcolor{mygreen}{$\uparrow$4.2})
\\

\textsc{Qwen3.8-2.4T-A95B}
& \textcolor{mygray}{\textbf{90.3}}
& 88.7 & 85.9
& 90.2 (\textcolor{mygreen}{$\uparrow$1.5})
\\

\midrule
\end{tabular}
}

\vspace{-4mm}
\label{tab:overall_comparison_large}
}}
\newcommand{\WeightKVAblationTable}{{
\begin{table*}[t]
\centering
\captionof{table}{
Ablation of isolated NVFP4 weight and NVFP4 KV on Qwen3.5-35B-A3B under reasoning RL tasks. $\dagger$ marks results quoted from QUADS~\citep{DBLP:journals/corr/abs-2607-15810}. The best performance is marked in bold. The performance gain compared to baselines is marked in green inside bracket.
}
\resizebox{\textwidth}{!}{%
\begin{tabular}{l|l|cccc|c}
\midrule

\textsc{Rollout Config}
& \textsc{Algo.}
& \textsc{LiveCodeBench}
& \textsc{AIME24}
& \textsc{AIME25}
& \textsc{HMMT25}
& \textsc{Average}
\\
\midrule

\textcolor{mygray}{BF16 W/A + BF16 KV}
& \textcolor{mygray}{Vanilla}
& \textcolor{mygray}{67.1}
& \textcolor{mygray}{\textbf{83.8}}
& \textcolor{mygray}{81.3}
& \textcolor{mygray}{67.5}
& \textcolor{mygray}{74.9}
\\
\midrule

\multirow{3}{*}{NVFP4 W/A + BF16 KV}
& QAT$^{\dagger}$
& 63.5
& 81.7
& 78.8
& 60.4
& 71.1
\\

& QUADS$^{\dagger}$
& 64.8
& 83.3
& 80.4
& 62.9
& 72.9
\\
\cmidrule{2-7}

& \textbf{\sysname}
& 67.1 (\textcolor{mygreen}{$\uparrow$2.3})
& 83.1
& \textbf{82.1} (\textcolor{mygreen}{$\uparrow$1.7})
& \textbf{69.4} (\textcolor{mygreen}{$\uparrow$6.5})
& \textbf{75.4} (\textcolor{mygreen}{$\uparrow$2.5})
\\

\midrule

\multirow{2}{*}{BF16 W/A + NVFP4 KV}
& QAT
& 65.3
& 82.3
& 79.2
& 65.8
& 73.2
\\
\cmidrule{2-7}

& \textbf{\sysname}
& \textbf{68.1} (\textcolor{mygreen}{$\uparrow$2.8})
& 83.0 (\textcolor{mygreen}{$\uparrow$0.7})
& 80.7 (\textcolor{mygreen}{$\uparrow$1.5})
& 67.3 (\textcolor{mygreen}{$\uparrow$1.5})
& 74.8 (\textcolor{mygreen}{$\uparrow$1.6})
\\

\midrule
\end{tabular}
}
\label{tab:weight_kv_ablation}
\end{table*}
}}

\newcommand{\PrecisionTable}{{
\begin{table}[t]
\centering
\captionof{table}{
Performance ($\uparrow$) comparison of stabilizing algorithms with different MXFP4 rollout configurations on Qwen3.5-35B-A3B under the reasoning RL task. The best performance is marked in bold. The performance gain compared to the QAT baseline is marked in green inside brackets.
}
\resizebox{1\textwidth}{!}{%
\begin{tabular}{l|l|cccc|c}
\midrule

\textsc{Rollout Config}
& \textsc{Algo.}
& \textsc{LiveCodeBench}
& \textsc{AIME24}
& \textsc{AIME25}
& \textsc{HMMT25}
& \textsc{Average}
\\
\midrule

\textcolor{mygray}{BF16 W/A + BF16 KV}
& \textcolor{mygray}{Vanilla}
& \textcolor{mygray}{\textbf{67.1}}
& \textcolor{mygray}{\textbf{83.8}}
& \textcolor{mygray}{\textbf{81.3}}
& \textcolor{mygray}{67.5}
& \textcolor{mygray}{74.9}
\\

\midrule

\multirow{2}{*}{W4A8 + MXFP4 KV}
& QAT
& 64.9
& 75.4
& 76.9
& 58.8
& 69.0
\\
\cmidrule{2-7}

& \textbf{\sysname}
& \textbf{67.1} (\textcolor{mygreen}{$\uparrow$2.2})
& 83.3 (\textcolor{mygreen}{$\uparrow$7.9})
& 81.2 (\textcolor{mygreen}{$\uparrow$4.3})
& \textbf{68.7} (\textcolor{mygreen}{$\uparrow$9.9})
& \textbf{75.1} (\textcolor{mygreen}{$\uparrow$6.1})
\\

\midrule

\multirow{2}{*}{W4A4 + MXFP4 KV}
& QAT
& 60.2
& 74.8
& 73.2
& 60.7
& 67.2
\\
\cmidrule{2-7}

& \textbf{\sysname}
& 65.2 (\textcolor{mygreen}{$\uparrow$5.0})
& 82.0 (\textcolor{mygreen}{$\uparrow$7.2})
& 79.8 (\textcolor{mygreen}{$\uparrow$6.6})
& 67.1 (\textcolor{mygreen}{$\uparrow$6.4})
& 73.5 (\textcolor{mygreen}{$\uparrow$6.3})
\\

\midrule
\end{tabular}
}
\label{tab:precision}
\end{table}
}}

\newcommand{\SensitivityTable}{{
\begin{table}[t]
\centering
\captionof{table}{
Modular sensitivity study of \sysname with FP4 MoE and FP4 KV on Qwen3.5-35B-A3B under reasoning RL tasks.
\sysname \texttt{(R-1bit-L20)} is the default config.
The best performance is marked in bold.
The performance gain compared to the QUADS baseline is marked in green inside brackets.
}
\resizebox{1\textwidth}{!}{%
\begin{tabular}{l|l|cccc|c}
\midrule
\textsc{Rollout Config}
& \textsc{Algo.}
& \textsc{LiveCodeBench}
& \textsc{AIME24}
& \textsc{AIME25}
& \textsc{HMMT25}
& \textsc{Average}
\\
\midrule

\textcolor{mygray}{BF16 W/A + BF16 KV}
& \textcolor{mygray}{Vanilla}
& \textcolor{mygray}{\textbf{67.1}}
& \textcolor{mygray}{83.8}
& \textcolor{mygray}{\textbf{81.3}}
& \textcolor{mygray}{67.5}
& \textcolor{mygray}{74.9}
\\
\midrule

\multirow{8}{*}{FP4 W/A + FP4 KV}
& QUADS
& 60.4
& 80.5
& 75.4
& 59.0
& 68.8
\\
\cmidrule{2-7}

& \textbf{\sysname \texttt{(R-4bit-L40)}}
& 67.0 (\textcolor{mygreen}{$\uparrow$6.6})
& 86.1 (\textcolor{mygreen}{$\uparrow$5.6})
& 80.1 (\textcolor{mygreen}{$\uparrow$4.7})
& \textbf{70.5} (\textcolor{mygreen}{$\uparrow$11.5})
& \textbf{75.9} (\textcolor{mygreen}{$\uparrow$7.1})
\\

& \textbf{\sysname \texttt{(R-3bit-L40)}}
& 66.8 (\textcolor{mygreen}{$\uparrow$6.4})
& 85.9 (\textcolor{mygreen}{$\uparrow$5.4})
& 81.2 (\textcolor{mygreen}{$\uparrow$5.8})
& 66.8 (\textcolor{mygreen}{$\uparrow$7.8})
& 75.2 (\textcolor{mygreen}{$\uparrow$6.4})
\\

& \textbf{\sysname \texttt{(R-2bit-L40)}}
& \textbf{67.1} (\textcolor{mygreen}{$\uparrow$6.7})
& 86.1 (\textcolor{mygreen}{$\uparrow$5.6})
& 80.0 (\textcolor{mygreen}{$\uparrow$4.6})
& 67.2 (\textcolor{mygreen}{$\uparrow$8.2})
& 75.1 (\textcolor{mygreen}{$\uparrow$6.3})
\\

& \textbf{\sysname \texttt{(R-1bit-L40)}}
& 66.4 (\textcolor{mygreen}{$\uparrow$6.0})
& \textbf{86.3} (\textcolor{mygreen}{$\uparrow$5.8})
& 80.5 (\textcolor{mygreen}{$\uparrow$5.1})
& 70.0 (\textcolor{mygreen}{$\uparrow$11.0})
& 75.8 (\textcolor{mygreen}{$\uparrow$7.0})
\\
\cmidrule{2-7}

& \textbf{\sysname \texttt{(R-1bit-L20)}}
& 66.4 (\textcolor{mygreen}{$\uparrow$6.0}) 
& \textbf{86.3} (\textcolor{mygreen}{$\uparrow$5.8}) 
& 78.5 (\textcolor{mygreen}{$\uparrow$3.1}) 
& 70.0 (\textcolor{mygreen}{$\uparrow$11.0}) 
& 75.3 (\textcolor{mygreen}{$\uparrow$6.5}) 
\\

& \textbf{\sysname \texttt{(R-1bit-L10)}}
& 66.1 (\textcolor{mygreen}{$\uparrow$5.7})
& 82.7 (\textcolor{mygreen}{$\uparrow$2.2})
& \textbf{81.3} (\textcolor{mygreen}{$\uparrow$5.9})
& 66.4 (\textcolor{mygreen}{$\uparrow$7.4})
& 74.1 (\textcolor{mygreen}{$\uparrow$5.3})
\\

& \textbf{\sysname \texttt{(R-1bit-L5)}}
& 65.8 (\textcolor{mygreen}{$\uparrow$5.4})
& 81.2 (\textcolor{mygreen}{$\uparrow$0.7})
& 79.7 (\textcolor{mygreen}{$\uparrow$4.3})
& 66.9 (\textcolor{mygreen}{$\uparrow$7.9})
& 73.4 (\textcolor{mygreen}{$\uparrow$4.6})
\\

\midrule
\end{tabular}
}
\label{tab:sensitivity}
\end{table}
}}

\newcommand{\PTQTable}{{
\centering
\captionof{table}{
Final FP4 policy performance ($\uparrow$) under different RL training and quantization regimes on Qwen3.5-35B-A3B.
For PTQ baselines, RL training is completed with BF16 rollout before applying post-hoc FP4 quantization.
In contrast, \sysname uses joint FP4 weight/activation and FP4 KV-cache rollout throughout RL training.
}
\vspace{-2mm}
\resizebox{\linewidth}{!}{%
\begin{tabular}{l|cccc|c}
\midrule

\textsc{Method}
& \textsc{LiveCodeBench}
& \textsc{AIME24}
& \textsc{AIME25}
& \textsc{HMMT25}
& \textsc{Average}
\\
\midrule

\textcolor{mygray}{BF16}
& \textcolor{mygray}{67.1}
& \textcolor{mygray}{83.8}
& \textcolor{mygray}{81.3}
& \textcolor{mygray}{67.5}
& \textcolor{mygray}{74.9}
\\
\midrule

Vanilla NVFP4
& 63.9 
& 75.7
& 78.4
& 63.4
& 70.4
\\

4over6
& 64.1
& 83.8 
& 76.7
& 59.2
& 71.0
\\

H-Scale
& 65.5 
& 82.0
& 77.8
& 60.1
& 71.4
\\
\midrule

\textbf{\sysname}
& \textbf{66.4} (\textcolor{mygreen}{$\uparrow$ 0.9})
& \textbf{86.3} (\textcolor{mygreen}{$\uparrow$ 2.5})
& \textbf{78.5} (\textcolor{mygreen}{$\uparrow$ 0.1})
& \textbf{70.0} (\textcolor{mygreen}{$\uparrow$ 6.6})
& \textbf{75.3} (\textcolor{mygreen}{$\uparrow$ 3.9})
\\
\midrule
\end{tabular}
}
\vspace{-4mm}
\label{tab:quant_weight}
}}

\newcommand{\SCTable}{{
\centering
\captionof{table}{
Performance ($\uparrow$) of SC and \sysname on Qwen3.5-35B-A3B under joint NVFP4 W/A/KV rollout.
SC is selected by the highest average across the four benchmarks through 400 steps. The best score in each column is marked in bold.}
\vspace{-2mm}
\setlength{\tabcolsep}{3pt}
\resizebox{\linewidth}{!}{%
\begin{tabular}{l|cccc|c}
\toprule
\textsc{Method}
& \textsc{LiveCodeBench}
& \textsc{AIME24}
& \textsc{AIME25}
& \textsc{HMMT25}
& \textsc{Average}
\\
\midrule
\textcolor{mygray}{BF16}
& \textcolor{mygray}{\textbf{67.1}}
& \textcolor{mygray}{83.8}
& \textcolor{mygray}{\textbf{81.3}}
& \textcolor{mygray}{67.5}
& \textcolor{mygray}{74.9}
\\\midrule

TIS+SC
& 66.7
& 81.7
& \textbf{81.3}
& 66.7
& 74.1
\\\midrule
\textbf{\sysname}
& 66.4
& \textbf{86.3} (\textcolor{mygreen}{$\uparrow$4.6}) 
& 78.5
& \textbf{70.0} (\textcolor{mygreen}{$\uparrow$3.3}) 
& \textbf{75.3} (\textcolor{mygreen}{$\uparrow$1.2}) 
\\
\bottomrule
\end{tabular}
}
\vspace{-4mm}
\label{tab:sc}
}}

\usepackage{microtype}
\usepackage{graphicx}
\usepackage{subfigure}
\usepackage{caption}
\usepackage{xspace}
\usepackage{pifont}
\usepackage{makecell}
\usepackage{bm}

\usepackage{booktabs} 
\usepackage{wrapfig}

\PassOptionsToPackage{numbers, compress}{natbib}
\usepackage{colm2024_conference}

\usepackage[utf8]{inputenc} 
\usepackage[T1]{fontenc}    
\usepackage{hyperref}       
\usepackage{url}            
\usepackage{amsfonts}       
\usepackage{nicefrac}       
\usepackage{xcolor}         

\usepackage{amsmath}
\usepackage{amssymb}
\usepackage{mathtools}
\usepackage{amsthm}
\usepackage{multirow}
\usepackage{graphicx}
\usepackage{lscape}
\usepackage{algorithm}
\usepackage{algpseudocode}
\usepackage[capitalize,noabbrev]{cleveref}

\theoremstyle{plain}

\theoremstyle{definition}

\theoremstyle{remark}

\newcommand{\sysname}{\texttt{TRACE}\xspace}

\usepackage[textsize=tiny]{todonotes}
\newcommand{\blfootnote}[1]{%
  \begingroup
  \renewcommand{\thefootnote}{}%
  \footnotetext{#1}%
  \endgroup
}

\title{TRACE: Rollout-Guided Quantization-Aware Training for FP4 Reinforcement Learning of MoE Language Models}
\author{%
\textbf{Xin Wang\textsuperscript{1,2,$*$,$\ddagger$} \quad
Hao Yu\textsuperscript{1,$*$} \quad
Zhengyang Zhuge\textsuperscript{1,$*$} \quad
Bochao Mao\textsuperscript{1} \\[0.1em]
Zheng Li\textsuperscript{1} \quad
Junda Feng\textsuperscript{1} \quad
Yuyan Luo\textsuperscript{1} \quad
Yi Zhang\textsuperscript{1} \\[0.1em]
Yizhong Cao\textsuperscript{1,$\dagger$} \quad
Mi Zhang\textsuperscript{2,$\dagger$} \quad
Dayiheng Liu\textsuperscript{1} \quad
Jianwei Zhang\textsuperscript{1,$\dagger$}}
\\[0.6em]
\normalfont
\textsuperscript{1}Alibaba Token Hub, Alibaba Group,
\quad
\textsuperscript{2}Ohio State University
}
\begin{document}

\definecolor{mygreen}{HTML}{009901}
\definecolor{myred}{HTML}{A52A2A}
\definecolor{mygray}{HTML}{9B9B9B}

\maketitle

\blfootnote{%
$^*$Equal contribution.}

\blfootnote{%
$^\dagger$Corresponding author.
}

\blfootnote{%
$^\ddagger$Work is done during internship at Alibaba Token Hub, Alibaba Group.
}

\begin{abstract}
Reinforcement learning (RL) for post-training large language models (LLMs) incurs substantial computation and memory overhead during rollout generation, which motivates low-precision rollout for efficient RL training.
However, existing FP4 RL methods suffer from a key limitation: they primarily optimize quantization accuracy on the training and rollout paths independently rather than directly reducing the discrepancy between the two quantized execution paths.
In this work, we propose \sysname (\textbf{T}rain-\textbf{R}ollout Quantization \textbf{A}lignment via \textbf{C}ompact Guidanc\textbf{E}), an FP4 quantization framework for RL training of Mixture-of-Experts (MoE) language models that addresses the limitation of existing FP4 RL methods.
\sysname incorporates rollout-guided quantization-aware training that uses rollout-side quantization outcomes to guide training-side FP4 rounding decisions, directly reducing train-rollout discrepancy.
Moreover, \sysname adopts an efficient quantization-information caching scheme that selectively retains mantissa and scale information from deeper layers to reduce the storage and communication overhead introduced by rollout guidance.
We evaluate \sysname on four large-scale MoE language models across reasoning, coding, and long-horizon RL tasks.
Our results demonstrate that \sysname enables joint FP4 weight/activation and FP4 KV-cache rollout with RL performance comparable to BF16 rollout, while achieving up to 5.4$\times$ rollout speedup and strong final FP4 performance compared with post-hoc FP4 quantization of BF16-trained policies.
\end{abstract}

\section{Introduction}
\label{sec:introduction}

Reinforcement learning (RL) has become an important post-training paradigm for improving the reasoning, coding, and other complex capabilities of Large Language Models (LLMs)~\citep{DBLP:journals/corr/abs-2303-18223,DBLP:journals/corr/abs-2509-08827}.
Despite its effectiveness, RL training incurs substantial computation and memory overhead due to the repeated generation of long trajectories during rollout~\citep{DBLP:journals/corr/abs-2506-06122}.
Low-precision rollout~\citep{DBLP:journals/corr/abs-2607-15810,anyscale_fp8_rl_skyrl,DBLP:journals/corr/abs-2602-13953,DBLP:journals/corr/abs-2601-14243,DBLP:journals/corr/abs-2510-11696} provides a promising solution by quantizing the policy model during generation to reduce rollout cost.
In particular, aggressive FP4 quantization (W4A4)  provides greater opportunities to accelerate rollout generation and reduce its memory footprint.
However, the coarse quantization space of FP4 also introduces large numerical discrepancies between the rollout and training execution paths, which can result in policy mismatch and destabilize RL training.
Such inconsistency is particularly challenging for Mixture-of-Experts (MoE) language models~\citep{DBLP:conf/iclr/ShazeerMMDLHD17}, where numerical differences may further interact with expert routing and amplify the mismatch between rollout and training policies~\citep{DBLP:journals/corr/abs-2512-01374}.
To mitigate this issue, several recent FP4 RL methods~\citep{DBLP:journals/corr/abs-2607-15810,DBLP:journals/corr/abs-2607-26515} improve the numerical consistency between low-precision rollout and training through quantization-aware training or quantization-error compensation.

Despite these efforts, existing FP4 RL methods suffer from a fundamental issue:
\textbf{Misalignment between quantization accuracy and train-rollout discrepancy.}
Existing methods primarily improve quantization accuracy with respect to the high-precision representation on the training and rollout paths independently.
However, minimizing the quantization error on each execution path does not directly minimize the discrepancy between the two quantized execution paths.
As a consequence, even when both paths achieve small quantization errors, their quantized values can still diverge substantially, leading to large policy mismatch, degraded RL performance, or even training collapse.

In this paper, we propose \sysname (\textbf{T}rain-\textbf{R}ollout Quantization \textbf{A}lignment via \textbf{C}ompact Guidanc\textbf{E}), an FP4 quantization framework for RL training of MoE language models.
\sysname differs from existing FP4 RL approaches in two key aspects.
\textbf{\ding{202} Rollout-Guided Quantization-Aware Training:}
\sysname directly reduces local quantized train-rollout discrepancy by using rollout-side quantization outcomes to guide training-side FP4 rounding decisions.
\textbf{\ding{203} Efficient Quantization-Information Caching:}
To reduce the storage and communication overhead introduced by rollout guidance during RL training, \sysname only retains the mantissa and scale information from selected deeper layers.

\begin{figure}[t]
    \centering
    \includegraphics[width=\linewidth]{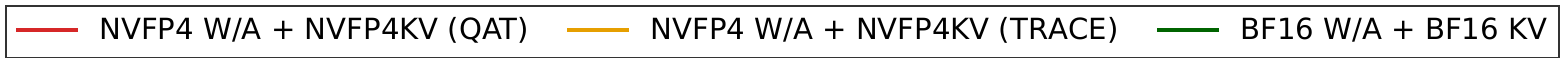}
    \subfigure[Train-Rollout Log-Prob Diff]{
        \centering
        \includegraphics[width=0.31\textwidth]{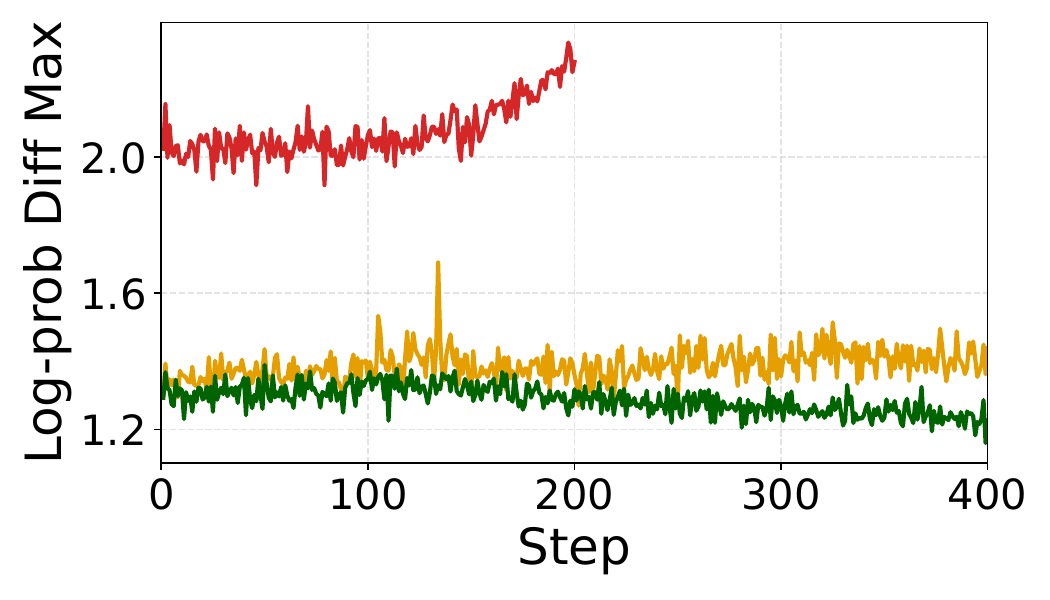}
        \label{fig:intro_logprob}
    }
    \hfill
    \subfigure[Test Score on Terminal-Bench]{
        \centering
        \includegraphics[width=0.31\textwidth]{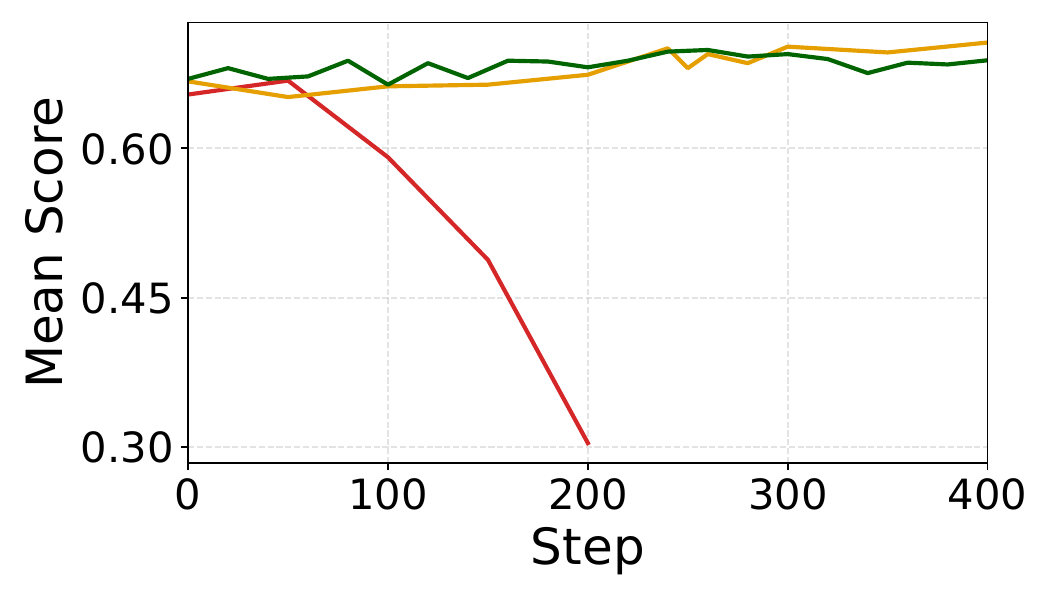}
        \label{fig:intro_validation}
    }
    \hfill
    \subfigure[Low-Precision Adaptation]{
        \centering
        \includegraphics[width=0.31\textwidth]{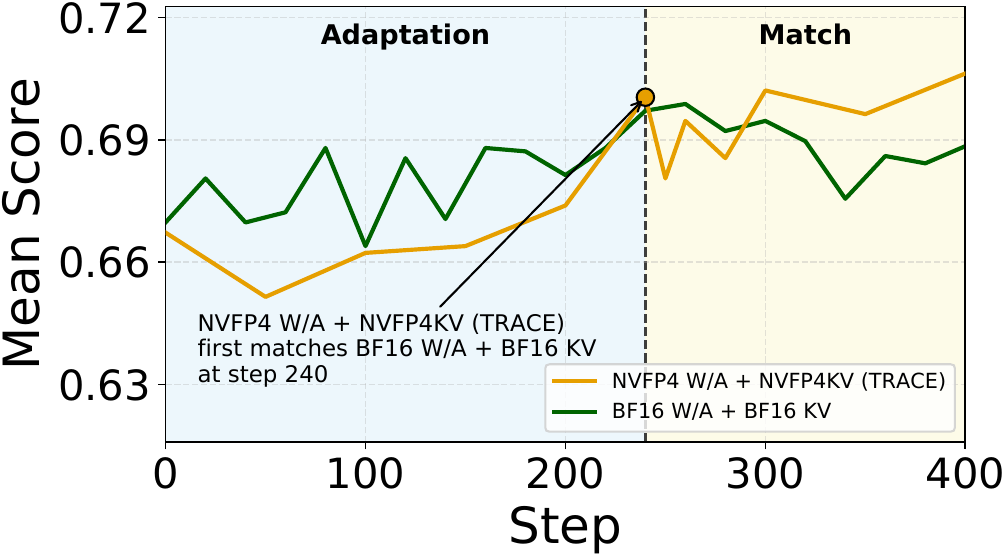}
        \label{fig:intro_adaptation}
    }

    \vspace{-2mm}
    \caption{
    Training dynamics of Qwen3.8-Flash-Next under FP4 rollout.
    Conventional FP4 QAT exhibits increasing train-rollout discrepancy and degraded test score on Terminal-Bench 2.1, while \sysname maintains stable alignment and enables the policy to progressively adapt to FP4 rollout.
    }
    \vspace{-2mm}
    \label{fig:intro_dynamics}
\end{figure}

Figure~\ref{fig:intro_dynamics} highlights the training dynamics under aggressive FP4 rollout (NVFP4 W/A + NVFP4 KV) with \sysname.
FP4 QAT exhibits an increasingly large train-rollout policy mismatch as training proceeds, resulting in a sharp degradation in test score on Terminal-Bench.
In contrast, \sysname effectively controls the train-rollout discrepancy and maintains stable RL training.
More interestingly, once the discrepancy is properly controlled, the policy progressively adapts to FP4 rollout: although \sysname initially suffers from quantization-induced performance degradation, 
the performance gap relative to BF16 rollout gradually diminishes during training and can eventually be closed or even reversed.

We compare \sysname with representative FP4 rollout baselines, including vanilla QAT, QaRL, QUADS and Score Centering, as well as post-hoc FP4 quantization methods, including vanilla NVFP4, 4over6, and H-Scale. To demonstrate the generalizability of \sysname, we evaluate it on four MoE language models---Qwen3.5-35B-A3B, Qwen3.5-122B-A10B, Qwen3.8-Flash-Next, and Qwen3.8-2.4T-A95B---across reasoning, coding, and long-horizon RL tasks.
We highlight three of our findings:
\begin{itemize}
    \item \sysname enables joint FP4 W/A and FP4 KV rollout while preserving BF16-level performance, maintaining training dynamics and final task performance comparable to BF16 rollout.

    \item \sysname retains the efficiency advantage of FP4 rollout with limited algorithm overhead, achieving up to $5.4\times$ rollout speedup over BF16 rollout.

    \item \sysname achieves stronger final FP4 task performance than pipelines that first train with BF16 rollout and apply post-hoc FP4 quantization.
\end{itemize}

\section{Related Work}
\label{sec:related_works}
\textbf{Reinforcement Learning for MoE Language Models:}
Modern LLM post-training increasingly relies on reinforcement learning (RL), with GRPO~\citep{DBLP:journals/corr/abs-2402-03300} being widely adopted for reasoning models. 
For Mixture-of-Experts (MoE) models~\citep{DBLP:conf/iclr/ShazeerMMDLHD17}, RL training is particularly sensitive to train--rollout inconsistency because small numerical differences can alter router scores and activate different experts~\citep{DBLP:journals/corr/abs-2512-01374}. 
R3~\citep{DBLP:journals/corr/abs-2510-11370} mitigates this routing mismatch by replaying rollout routing decisions during training, while PR2~\citep{DBLP:journals/corr/abs-2606-00395} further accounts for router evolution across policy updates. 
GSPO~\citep{DBLP:journals/corr/abs-2507-18071} instead improves stability through sequence-level importance ratios and clipping. 
Meanwhile, the large parameter capacity of MoE models introduces substantial memory, communication, and rollout costs, motivating efficient distributed training and low-precision rollout~\citep{DBLP:journals/corr/abs-2506-06122, DBLP:journals/corr/abs-2401-01923}.

\textbf{Low-Precision Reinforcement Learning:} 
Recent works have explored low-precision rollout to reduce the generation cost of RL for LLMs.
For MoE models, quantization is often applied only to routed experts while keeping other modules in higher precision~\citep{DBLP:journals/corr/abs-2607-15810,anyscale_fp8_rl_skyrl,DBLP:journals/corr/abs-2608-28113}.
Most prior methods focus on 8-bit precision.
For example, QuRL~\citep{DBLP:journals/corr/abs-2602-13953} reduces train--rollout mismatch through adaptive clipping and update-aware quantization, while Jet-RL~\citep{DBLP:journals/corr/abs-2601-14243} adopts a unified FP8 computation flow.
More recent works extend rollout quantization to 4-bit precision.
QeRL~\citep{DBLP:journals/corr/abs-2510-11696} mainly considers 4-bit weight quantization with higher-precision activations; QUADS~\citep{DBLP:journals/corr/abs-2607-15810} combines asymmetric QAT with rollout-side activation compensation; Rollout-ResQ~\citep{DBLP:journals/corr/abs-2607-26515} corrects FP4 rollout errors with sparse residuals; and QaRL~\citep{DBLP:conf/acl/GuWLLZLXWYLHG26} aligns training and rollout through matched low-precision kernels.
Orthogonally, Score Centering~\citep{marek2026score} stabilizes RL by correcting optimization drift induced by training--inference mismatch, without directly aligning the underlying low-precision computations.
However, even with matched execution or optimization-level correction, small activation differences can still cross FP4 rounding boundaries and produce different codewords.
In contrast, \sysname directly uses rollout-side FP4 outcomes to guide training-side rounding, explicitly targeting this rounding-outcome mismatch.
\begin{figure*}[t]
\centering
\includegraphics[width=1\textwidth]{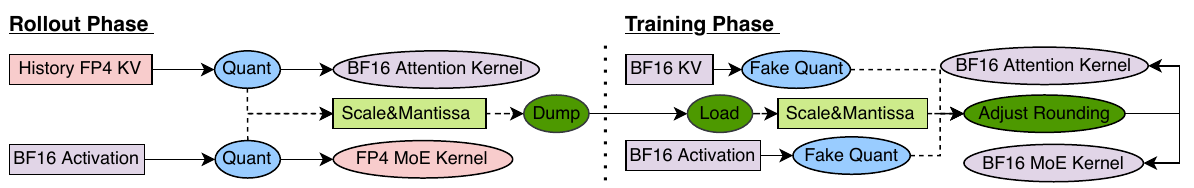}
\vspace{-3mm}
\caption{Overview of \sysname for FP4 RL (FP4 W/A + FP4 KV) Training of MoE Models
.}
\vspace{-3mm}
\label{fig:framework}
\end{figure*}

\section{\textbf{T}rain-\textbf{R}ollout Quantization \textbf{A}lignment via \textbf{C}ompact Guidanc\textbf{E} (TRACE)}
\label{sec:methodology}
\cref{fig:framework} provides an overview of \sysname. At a high level,  \sysname is an FP4 quantization framework for efficient RL training of MoE language models that directly aligns the quantized computation paths used for rollout generation and training. Specifically, during rollout generation, \sysname records the quantization outcomes of FP4 routed experts's activation and FP4 KV states. During the subsequent quantization-aware training (QAT) phase, the recorded rollout-side information is used to guide the corresponding training-side rounding decisions. To reduce the overhead of rollout guidance, \sysname further adopts an efficient quantization-information caching scheme. It retains only the mantissa and scale information from selected deeper layers to reduce storage and communication costs. In the following, we describe rollout-guided quantization-aware training and efficient quantization-information caching in detail.
The pseudocode of \sysname is provided in~\cref{appendix:pseudocode}.

\subsection{Rollout-Guided Quantization-Aware Training}
\label{subsec:rollout_guided_qat}

\textbf{Motivation:}
Low-precision rollout introduces numerical discrepancy between the training and rollout execution paths.
Simply combining FP4 quantization-aware training (QAT), where FP4 fake quantization is applied during the forward pass while the backward pass remains in BF16, with FP4 rollout generation can therefore lead to substantial policy divergence and degraded or even unstable RL training~\citep{DBLP:journals/corr/abs-2607-15810}. For paired training- and rollout-side activations corresponding to the same trajectory, token position, and quantization site, we characterize their quantized discrepancy as
\begin{equation}
    \mathcal{D}_{\mathrm{act}}
    =
    \left\|
    Q^{\mathrm{train}}_{\mathrm{FP4}}(X_{\mathrm{train}})
    -
    Q^{\mathrm{rollout}}_{\mathrm{FP4}}(X_{\mathrm{rollout}})
    \right\|_F ,
    \label{eq:act_discrepancy}
\end{equation}
where $X_{\mathrm{train}}$ and $X_{\mathrm{rollout}}$ denote the corresponding activations produced during training and rollout generation, respectively, and
$Q^{\mathrm{train}}_{\mathrm{FP4}}$ and
$Q^{\mathrm{rollout}}_{\mathrm{FP4}}$
denote the FP4 quantization procedures on the two execution paths.
We use $\mathcal{D}_{\mathrm{act}}$ to characterize local train--rollout discrepancy.

\begin{wrapfigure}{r}{0.6\textwidth}
  \centering
  \vspace{-7mm}
  \includegraphics[width=0.6\textwidth]{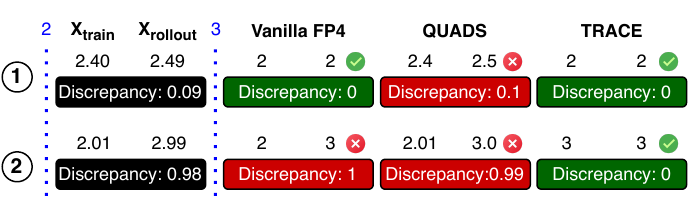}
  \caption{Two Examples of Train--Rollout Discrepancy under Different FP4 RL Methods.}
  \label{fig:rounding_example}
  \vspace{-3mm}
\end{wrapfigure}
Existing methods such as QUADS~\citep{DBLP:journals/corr/abs-2607-15810} and Rollout-ResQ~\citep{DBLP:journals/corr/abs-2607-26515} mitigate this discrepancy indirectly by improving the quantization accuracy of each execution path.
Specifically, they primarily reduce the quantization error with respect to the corresponding high-precision representation, e.g.,
$\mathcal{L}_{\mathrm{error}}(X)=\|X-Q(X)\|_F$.
However, minimizing the quantization errors of the training and rollout paths independently does not directly minimize their cross-path discrepancy.
To illustrate this issue, we provide two examples of QUADS in~\cref{fig:rounding_example}.
QUADS retains the BF16 training activation and compensates for the FP4 rollout quantization error using a residual $R(X)=X-Q(X)$.
In example \ding{202}, the training and rollout activations are $2.40$ and $2.49$, respectively.
Vanilla FP4 maps both values to $2$, yielding zero quantized discrepancy.
In contrast, QUADS retains $2.40$ on the training side and reconstructs the rollout activation as $2.50$, reducing the per-path quantization error but increasing the train--rollout discrepancy to $0.10$.
In example \ding{203}, QUADS similarly reduces the rollout-side quantization error but leaves a discrepancy of $0.99$, close to the discrepancy of $1.0$ under vanilla FP4.
These examples demonstrate that improving per-path quantization accuracy does not necessarily reduce train--rollout discrepancy.

\begin{figure*}[th]
\centering
\begin{minipage}[t]{0.58\textwidth}
\centering
\includegraphics[width=\linewidth]{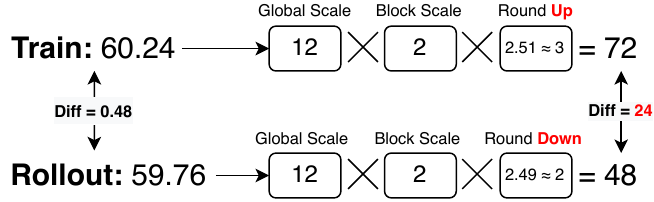}
{\small (a) Discrepancy Amplification by Inconsistent FP4 Rounding}
\end{minipage}
\hfill
\begin{minipage}[t]{0.4\textwidth}
\centering
\includegraphics[width=\linewidth]{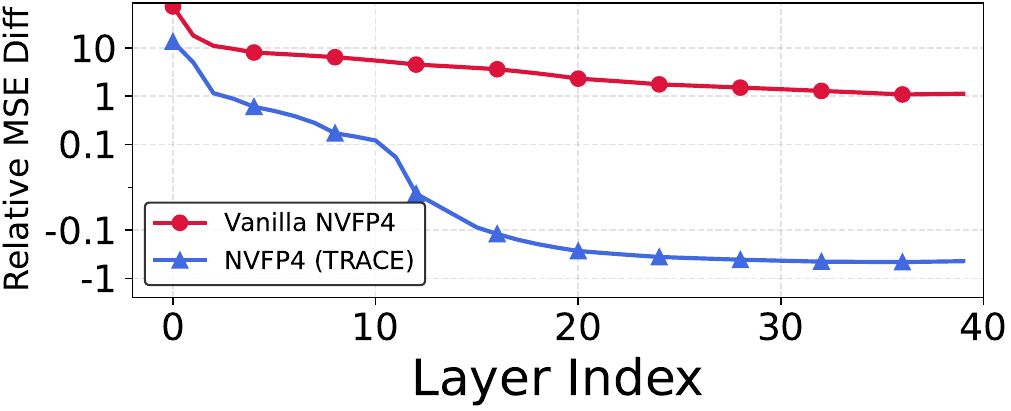}
{\small (b) Relative Excess Discrepancy}
\end{minipage}
\caption{Problem Analysis of FP4 RL Methods on Qwen3.5-35B-A3B.}
\label{fig:rounding_analysis}
\end{figure*}

\textbf{Key Design:}
The key idea of \sysname is to reduce the additional train--rollout discrepancy introduced by inconsistent FP4 rounding.
A key source of this discrepancy is that small differences between the BF16 training and rollout activations can be substantially amplified when they are rounded to different FP4 codewords.
As illustrated in~\cref{fig:rounding_analysis}(a), the original training and rollout activations, $60.24$ and $59.76$, differ by only $0.48$.
After applying the same global and block scales, however, their normalized values, $2.51$ and $2.49$, fall on opposite sides of the rounding boundary and are rounded to $3$ and $2$, respectively.
Consequently, the discrepancy is amplified from $0.48$ in BF16 to $24$ after FP4 quantization.
We further quantify this quantization-induced discrepancy in~\cref{fig:rounding_analysis}(b) using the relative excess train--rollout MSE,
$(D_{\mathrm{FP4}}-D_{\mathrm{BF16}})/D_{\mathrm{BF16}}$,
where $D_{\mathrm{FP4}}$ and $D_{\mathrm{BF16}}$ denote the train--rollout activation MSE under FP4 and BF16 computation, respectively.
As shown, vanilla NVFP4 introduces substantial additional discrepancy across model layers.

Based on this, \sysname uses the rollout-side quantization outcome to guide training-side rounding.
For each captured training-side activation, let
$\{q_{-},q_{+}\}$ denote its two neighboring normalized FP4 E2M1 codewords under the rollout-side scale, and let
$q_{\mathrm{rollout}}$ denote the exact FP4 codeword produced by the corresponding rollout-side activation.
Instead of applying standard round-to-nearest (RTN), \sysname selects
\begin{equation}
    q_{\mathrm{TRACE}}
    =
    \arg\min_{q\in\{q_{-},q_{+}\}}
    \left|q-q_{\mathrm{rollout}}\right|.
    \label{eq:trace_local}
\end{equation}
Let $q_{\mathrm{RTN}}\in\{q_{-},q_{+}\}$ denote the codeword selected by standard RTN under the same scale and candidate set.
By construction,
\begin{equation}
    \left|
    q_{\mathrm{TRACE}}-q_{\mathrm{rollout}}
    \right|
    \leq
    \left|
    q_{\mathrm{RTN}}-q_{\mathrm{rollout}}
    \right|.
    \label{eq:trace_guarantee}
\end{equation}
Therefore, under the full-information formulation, relative to RTN, \sysname does not increase the local quantized train--rollout discrepancy at each captured site under the same candidate set.

This local guarantee does not imply monotonic reduction of the full-network discrepancy in~\cref{eq:act_discrepancy} or of policy-level divergence after downstream nonlinear transformations.
Rather, \sysname reduces unnecessary amplification caused by inconsistent FP4 rounding.
As illustrated in~\cref{fig:rounding_example}, rollout-guided rounding eliminates the quantized discrepancy in both examples, while~\cref{fig:rounding_analysis}(b) shows substantially reduced train--rollout discrepancy across model layers. 
The formulation above assumes access to the exact rollout-side FP4 codeword.
In~\cref{subsec:mantissa_only_train_infer_communication}, we introduce a compressed caching scheme that approximates this guidance using only partial rollout-side code information.

\subsection{Mantissa-Only Train-Rollout Communication}
\label{subsec:mantissa_only_train_infer_communication}
\begin{figure*}[t]
\centering
\includegraphics[width=0.98\textwidth]{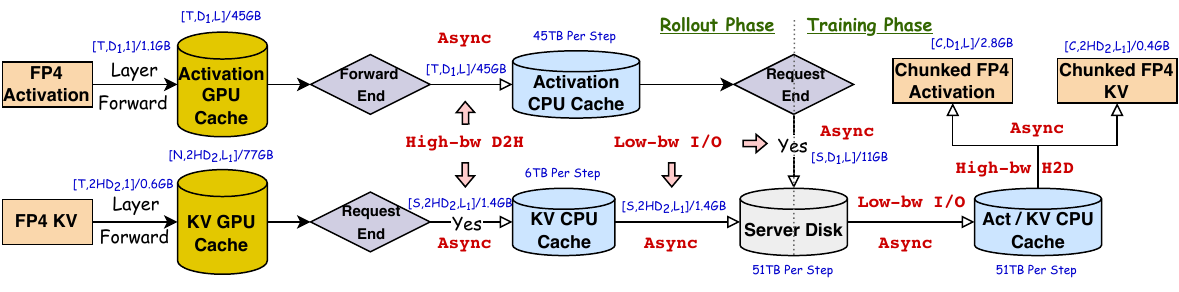}
\vspace{-3mm}
\caption{Data Communication Pipeline of Rollout-Side Quantization Information.}
\vspace{-3mm}
\label{fig:communication_overhead}
\end{figure*}
\textbf{Motivation:}
Although rollout-guided QAT effectively reduces train--rollout discrepancy, preserving complete rollout-side quantization information introduces substantial data-movement overhead.
As illustrated in~\cref{fig:communication_overhead}, the activation-side and KV-side quantization records follow different collection paths during rollout.
For activation guidance, the intermediate quantization information generated during each forward is first written into a temporary GPU buffer and asynchronously offloaded to a CPU-side cache after the forward completes. The temporary GPU buffer is reused by subsequent forwards.
In contrast, the quantized KV states are already retained in the persistent FP4 KV cache and therefore only need to be gathered and transferred to CPU memory when the corresponding request finishes.
After request completion, both activation and KV quantization records are serialized and persisted to storage.
Before QAT, the training engine reloads the records for the forward microbatches into CPU memory, after which the corresponding slices are transferred to GPUs.

The amount of data generated by this full-information pipeline can be extremely large.
For example, consider Qwen3.5-35B-A3B with hidden size $D_1=2048$, KV head dimension $D_2=256$, two KV heads, 40 MoE layers, 10 full-attention layers, and a maximum response length of 256K tokens.
Assuming 4.5 bits are retained per quantized value, one generated token produces approximately 45 KB of activation guidance and 5.6 KB of KV guidance, or about 50 KB in total.
With 4,096 trajectories in one RL step, this corresponds to up to approximately 51 TB of rollout-side quantization information per step. Such a data volume is difficult to hide through asynchronous communication alone.
For illustration, assuming an effective GPU-to-CPU bandwidth of 300 GB/s, transferring 51 TB requires approximately three minutes in aggregate.
In contrast, at a representative storage bandwidth of 5 GB/s, writing or reading the same amount of data would require roughly three hours if performed serially, far exceeding the wall-clock time of a typical RL step.
Therefore, even if GPU-to-CPU transfer can be efficiently overlapped with rollout computation, storage I/O can easily become the dominant bottleneck once the rollout-side quantization records are persisted between rollout and training.

\begin{wrapfigure}{r}{0.48\textwidth}
  \centering
  \includegraphics[width=0.48\textwidth]{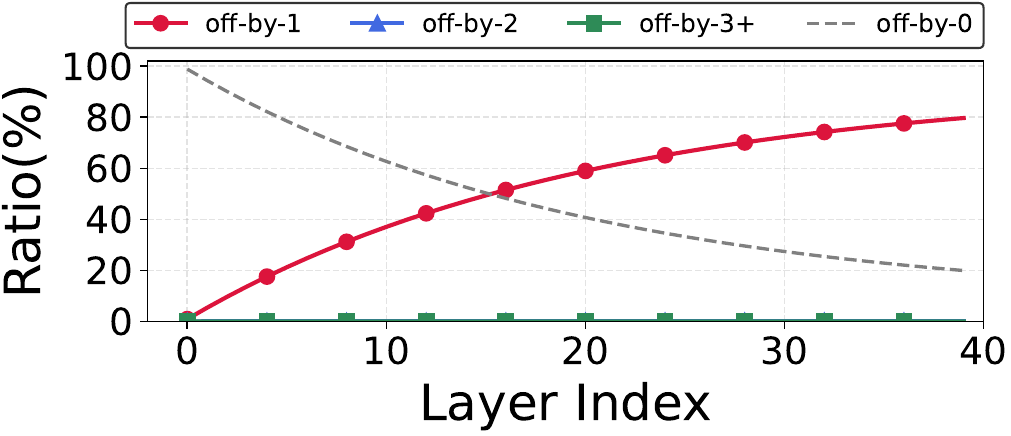}
  \caption{Difference in FP4 Codebook Entries Between Training and Rollout Activations.}
  \label{fig:mantissa_difference}
  \vspace{-3mm}
\end{wrapfigure}

\textbf{Key Design:} The key idea of \sysname is to communicate only the rollout-side quantization information necessary to determine the desired training-side rounding direction. As shown in~\cref{fig:mantissa_difference}, for over 99\% of the mismatched quantized values across layers, the training and rollout results differ by only one adjacent FP4 codebook entry. Because the two quantized values are overwhelmingly adjacent in the FP4 codebook, the training-side activation together with the rollout scale strongly constrains the candidate codewords. 
Consequently, a small amount of mantissa code information is typically sufficient to identify the desired neighboring rollout-side codeword. Therefore, communicating the complete quantized value is unnecessary. We further observe that rounding corrections toward lower FP4 codewords occur predominantly in deeper layers, indicating that rollout-side quantization information from these layers is more important for train-rollout alignment. Based on these observations, \sysname communicates only the mantissa and scale information from the latter half of the model layers. During rollout generation, this compact information is cached and transferred to the training engine, where it is used to reconstruct a compact rollout-side reference and guide training-side quantization.
This mantissa-only train-rollout communication substantially reduces the storage and communication overhead while preserving the effectiveness of \sysname.

\section{Experiment Analysis}
\label{sec:experiments}

\textbf{Baselines.}
We compare \sysname against three groups of methods:
(1) NVFP4 rollout methods, including vanilla NVFP4, NVFP4 with quantization-aware training (QAT), QaRL~\citep{DBLP:conf/acl/GuWLLZLXWYLHG26}, QUADS~\citep{DBLP:journals/corr/abs-2607-15810}~(Section~\ref{subsec:overall_performance}), and Score Centering~\citep{marek2026score} (Section~\ref{subsec:Comparison_with_sc});
(2) MXFP4 rollout methods, including vanilla and QAT variants under W4A8 and W4A4 settings~(Section~\ref{subsec:ablation_study}); and
(3) NVFP4 post-training quantization methods, including vanilla NVFP4, 4over6~\citep{DBLP:journals/corr/abs-2512-02010}, and H-Scale~\citep{DBLP:journals/corr/abs-2608-28113}~(Section~\ref{subsection:Comparison_with_ptq}).

\textbf{Models and Datasets.}
We evaluate \sysname on four MoE language models: Qwen3.5-35B-A3B and Qwen3.5-122B-A10B~\citep{qwen35blog}, Qwen3.8-Flash-Next~\citep{DBLP:journals/corr/abs-2608-30320} (125B total / 6B activated parameters), and Qwen3.8-2.4T-A95B~\citep{qwen38}. 
We consider three RL settings: reasoning, coding, and long-horizon tasks.
For reasoning RL, we train Qwen3.5-35B-A3B and evaluate it on LiveCodeBench~v6~\citep{DBLP:conf/iclr/JainHGLYZWSSS25}, AIME~2024/2025~\citep{aime24,aime25}, and HMMT~2025~\citep{balunovic2025matharena}.
For coding RL, we train Qwen3.5-122B-A10B and Qwen3.8-Flash-Next, and evaluate them on DeepSWE~v1.1~\citep{DBLP:journals/corr/abs-2607-07946} and Terminal-Bench~2.1~\citep{DBLP:journals/corr/abs-2601-11868}, respectively. For long-horizon RL, we train Qwen3.8-2.4T-A95B and evaluate it on GDPval~\citep{DBLP:journals/corr/abs-2510-04374}.
We report sampled average pass@1 over 8 responses for AIME24, AIME25, and HMMT25, and over 10 responses for LiveCodeBench.
For Terminal-Bench, DeepSWE, and GDPval, we report scores averaged over 10, 4, and 3 independent runs per task, respectively.

\textbf{Implementation Details.}
For fair comparison, all methods use the same versions of VeRL~\citep{DBLP:conf/eurosys/ShengZYWZZPL025}, Megatron~\citep{DBLP:journals/corr/abs-1909-08053}, and SGLang~\citep{DBLP:conf/nips/ZhengYXS0YCKSGB24}, where VeRL coordinates asynchronous RL training, Megatron serves as the training backend, and SGLang performs rollout generation. We use disagg architecture to deploy the trainer and rollout engines.
We adopt GRPO~\citep{DBLP:journals/corr/abs-2402-03300} for policy optimization and apply R3~\citep{DBLP:journals/corr/abs-2510-11370} to replay rollout-side expert-routing decisions during training.
We additionally use token-level importance ratios between the training and rollout policies~\citep{DBLP:journals/corr/abs-2512-01374}, with the clipping upper bound set to 5, and set \texttt{max\_version\_diff=3} to bound policy staleness introduced by asynchronous training.
For rollout-guided quantization, rollout-side quantization records are paired with the corresponding training-side activations by trajectory, token position, and quantization site.
Since the training policy can be newer than the policy that generated the rollout, \sysname does not eliminate the underlying activation difference caused by policy staleness; instead, it reduces the additional discrepancy introduced when such differences are amplified by inconsistent FP4 rounding.
Unless otherwise specified, we quantize only the routed experts in MoE layers and the KV cache in softmax-attention layers during rollout, while keeping all other modules in BF16.
For Qwen3.5-35B-A3B, Qwen3.5-122B-A10B, Qwen3.8-Flash-Next, and Qwen3.8-2.4T-A95B, each RL step samples $256{\times}16$, $128{\times}16$, $64{\times}16$, and $64{\times}8$ trajectories, 
respectively.
We set the maximum response length to 32K tokens for reasoning RL and the maximum context length to 256K tokens for coding and long-horizon RL.
Qwen3.5-35B-A3B and Qwen3.8-Flash-Next are trained for 400 RL steps, while Qwen3.5-122B-A10B and Qwen3.8-2.4T-A95B are trained for 50 steps.

\subsection{Overall Comparison}
\label{subsec:overall_performance}

We first evaluate the performance of \sysname under joint FP4 weight and KV cache quantization during rollout from three aspects:
(1) performance across different benchmarks on the reasoning RL task,
(2) performance on larger-scale MoE models under coding and long-horizon RL tasks, and
(3) performance evolution throughout RL training.
We compare \sysname with baselines under the same FP4 rollout configuration and use BF16 rollout as the high-precision reference.

\OverallComparisonTableSmall

\begin{wrapfigure}{r}{0.55\textwidth} 
  \centering
  \vspace{-5mm}
  \OverallComparisonTableLarge
\end{wrapfigure}

\textbf{Performance on Different Benchmarks.}
We evaluate \sysname on Qwen3.5-35B-A3B across reasoning benchmarks.
As shown in \Cref{tab:overall_comparison_35b}, \sysname consistently outperforms QAT, QaRL, and QUADS across all four benchmarks.
In particular, \sysname improves the average score from 68.8 to 75.3.
The improvement is especially significant on HMMT25, where the score increases from 59.0 to 70.0.
More importantly, \sysname achieves an average score matching the BF16 rollout, demonstrating that \sysname can effectively recover the performance degradation introduced by joint FP4 weight and KV cache rollout while achieving performance comparable to BF16 rollout.

\textbf{Performance on Larger-Scale Models.}
To examine the generalizability of \sysname to larger-scale MoE models, we further evaluate it on Qwen3.5-122B-A10B, Qwen3.8-Flash-Next, and Qwen3.8-2.4T-A95B under coding and long-horizon RL settings.
As shown in~\cref{tab:overall_comparison_large}, \sysname consistently outperforms QAT across all three models.
In particular, on Terminal-Bench trained with Qwen3.8-Flash-Next, \sysname improves the score from 66.4 to 70.6.
Moreover, \sysname matches BF16 rollout performance across all three models, achieving 33.0 vs.\ 33.4 on DeepSWE, 70.6 vs.\ 68.8 on Terminal-Bench, and 90.2 vs.\ 90.3 on GDPval.
These results demonstrate that the effectiveness of \sysname generalizes across different model scales and RL tasks while preserving the performance of high-precision BF16 rollout.

\begin{figure}[h]
 \vspace{-0mm}
     \begin{minipage}{0.98\textwidth}
     \centering
    \includegraphics[width=1\linewidth]{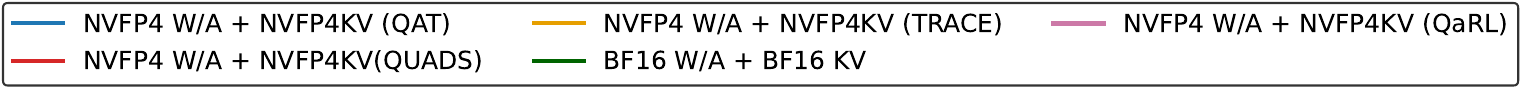}
     \end{minipage}
     \\
     \begin{minipage}{1\textwidth}
        \subfigure[Mean Reward]{
            \centering
            \includegraphics[width=0.31\textwidth]{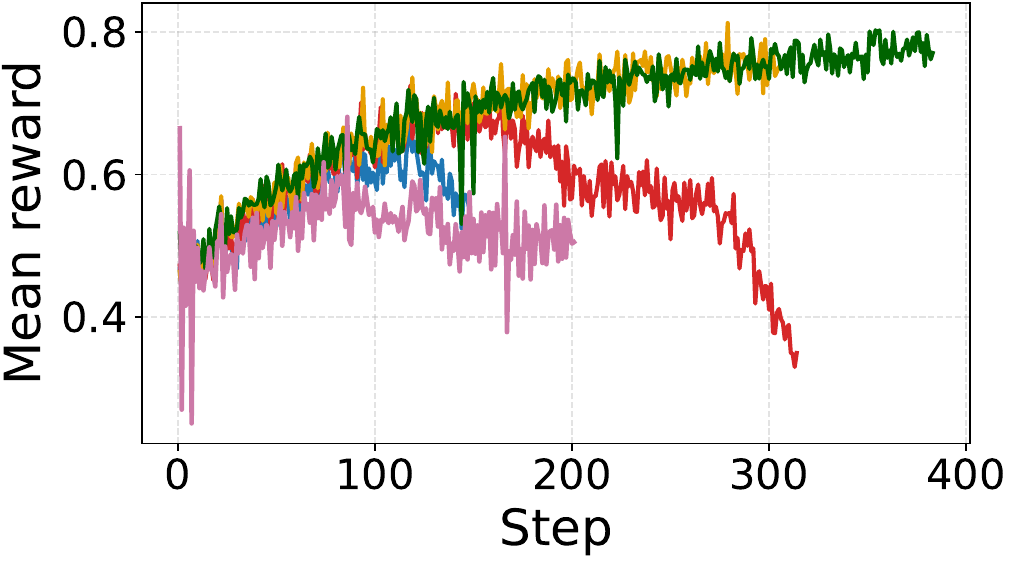}
            \label{fig:reward}
            }
        \subfigure[Mean Response Length]{
            \centering
            \includegraphics[width=0.31\textwidth]{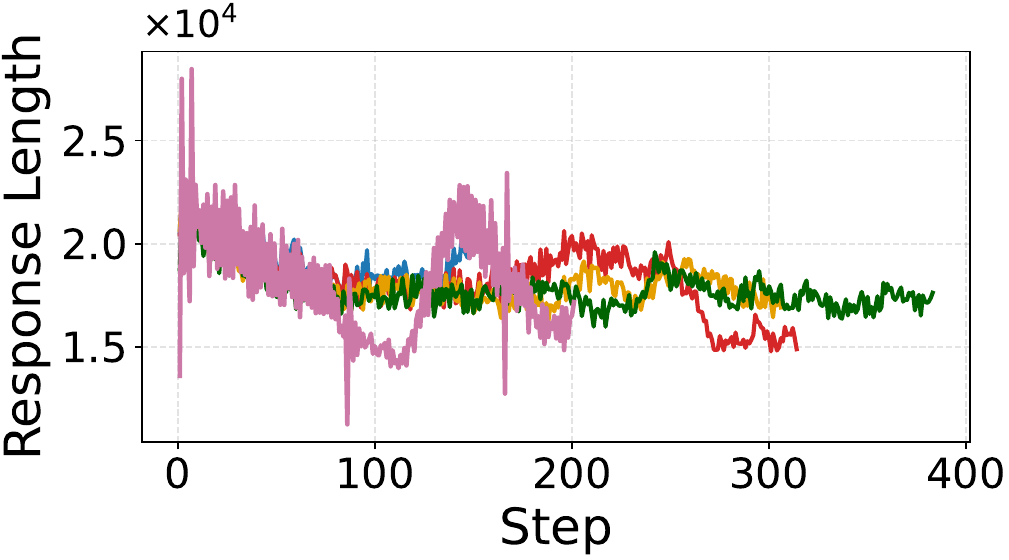}
            \label{fig:response_length}
            }
        \subfigure[Test Score on HMMT]{
            \centering
            \includegraphics[width=0.31\textwidth]{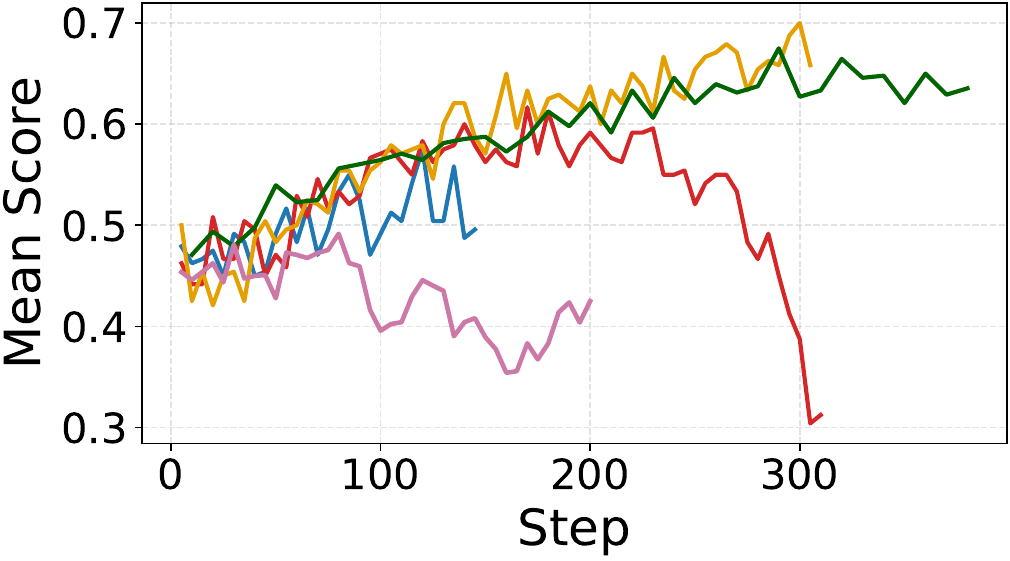}
            \label{fig:val_hmmt25}
            }
        \\
        \subfigure[Test Score on AIME24]{
            \centering
            \includegraphics[width=0.31\textwidth]{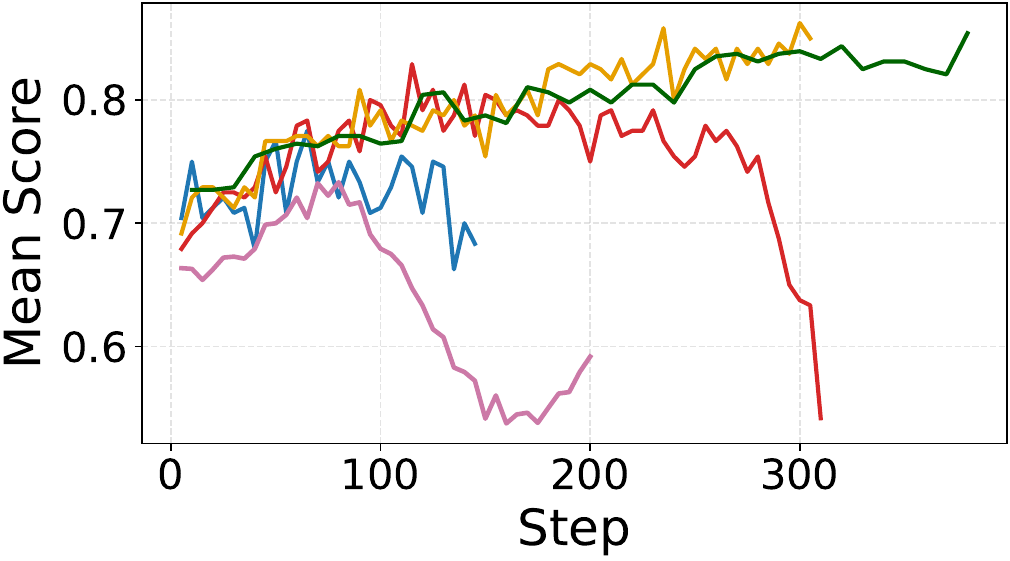}
            \label{fig:val_aime24}
            }
        \subfigure[Test Score on AIME25]{
            \centering
            \includegraphics[width=0.31\textwidth]{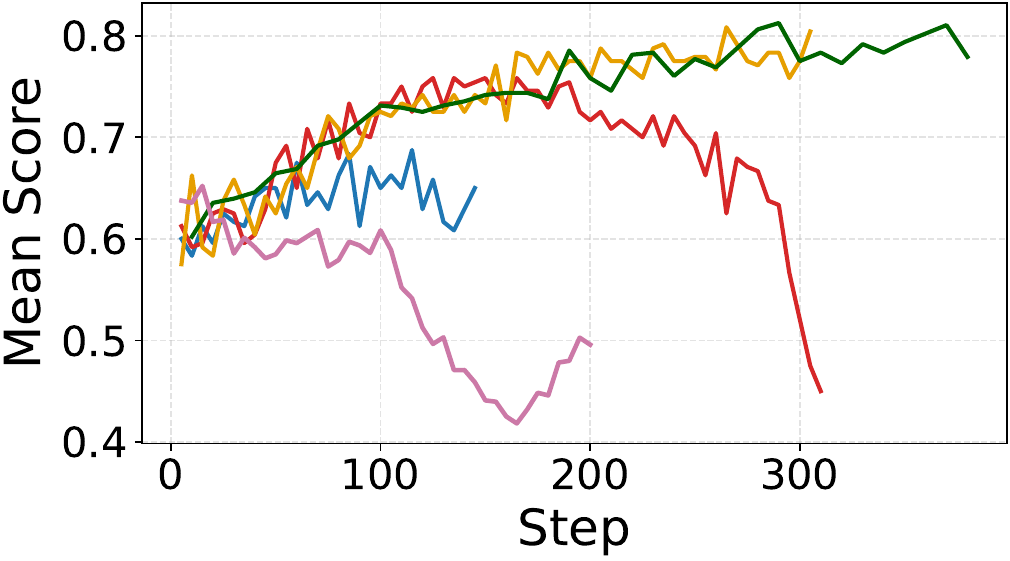}
            \label{fig:val_aime25}
            }
        \subfigure[Test Score on LCB.]{
            \centering
            \includegraphics[width=0.31\textwidth]{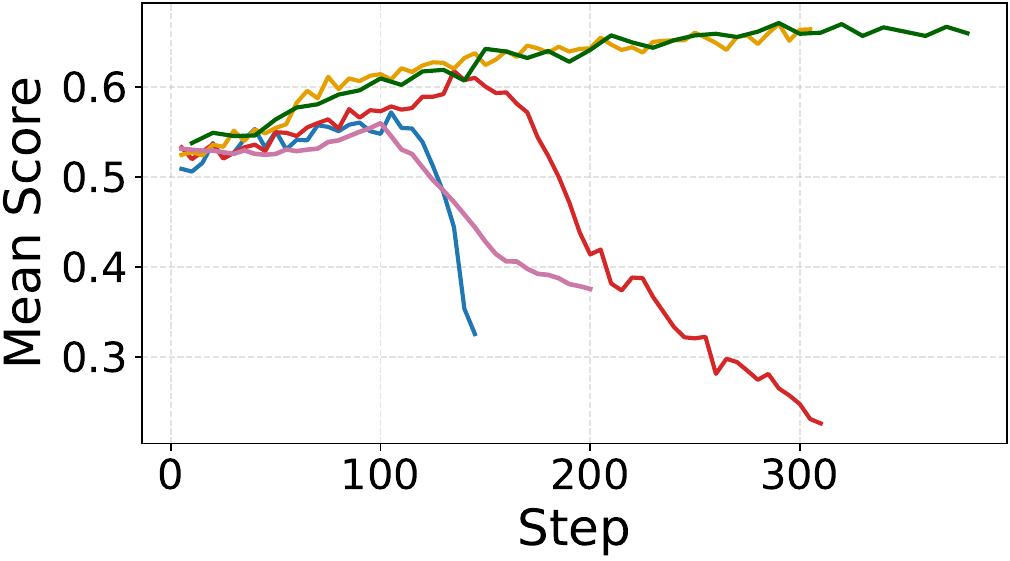}
            \label{fig:val_livecodebench}
            }
        \vspace{-2mm}
        \caption{Training dynamics of Qwen3.5-35B-A3B on reasoning RL tasks, including reward, response length, and test scores on HMMT, AIME24, AIME25, and LiveCodeBench.}
        \vspace{-2mm}
        \label{fig:training_curves}
    \end{minipage}
\end{figure}

\textbf{Training Dynamics Analysis.}
We examine the training dynamics of Qwen3.5-35B-A3B throughout RL training on reasoning tasks. The analysis on Qwen3.8-Flash-Next and Qwen3.8-2.4T-A95B are provided in~\cref{appendix:coding_training_curves} and~\cref{appendix:max_training_curves} respectively.
As shown in~\Cref{fig:training_curves}, we make two observations.
\begin{wrapfigure}{r}{0.46\textwidth} 
  \centering
  \includegraphics[width=\linewidth]{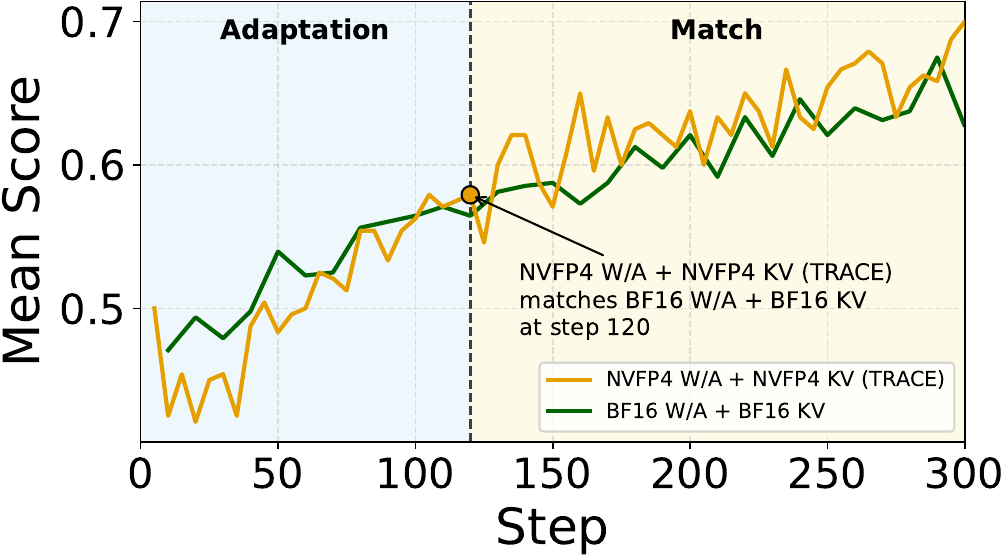} 
    \caption{Adaptation for NVFP4 (W/A+KV) with \sysname on HMMT25 during RL training.}
   \label{fig:training_curves_adaptation}
   \vspace{-4mm}
\end{wrapfigure}
(1) QAT, QaRL and QUADS exhibit increasingly large train--rollout discrepancy as training proceeds, which is eventually accompanied by clear signs of training instability.
In particular, the mean reward starts to decrease substantially at later training stages, while the test scores on all reasoning benchmarks degrade sharply.
In contrast, \sysname effectively controls the train--rollout discrepancy and maintains training dynamics close to BF16 rollout.
Both the training reward and downstream test performance remain stable and continue to improve throughout training.
These results further indicate that reducing train--rollout discrepancy is important for maintaining stable RL optimization under aggressive FP4 rollout.
(2) More importantly, we observe an interesting adaptation behavior under FP4 rollout with \sysname.
As illustrated in~\Cref{fig:training_curves_adaptation}, the test score of \sysname is initially lower than that of the BF16 rollout due to the accuracy degradation introduced by quantization.
However, \textbf{this performance gap gradually diminishes as RL training proceeds.}
After approximately 120 training steps, \sysname catches up with the BF16 reference and subsequently remains comparable, with higher point estimates at several later steps.
This trend suggests that \sysname does not merely preserve the initial model quality under FP4 execution.
Instead, once the train--rollout discrepancy is sufficiently controlled, the policy can progressively adapt to the low-precision execution environment during RL training, recovering the initial quantization-induced degradation and eventually matching the BF16 rollout.

\begin{figure}[h]
\vspace{-0mm}
\begin{minipage}{1\textwidth}
    \subfigure[Decoding Throughput]{
        \centering
        \includegraphics[width=0.52\textwidth]{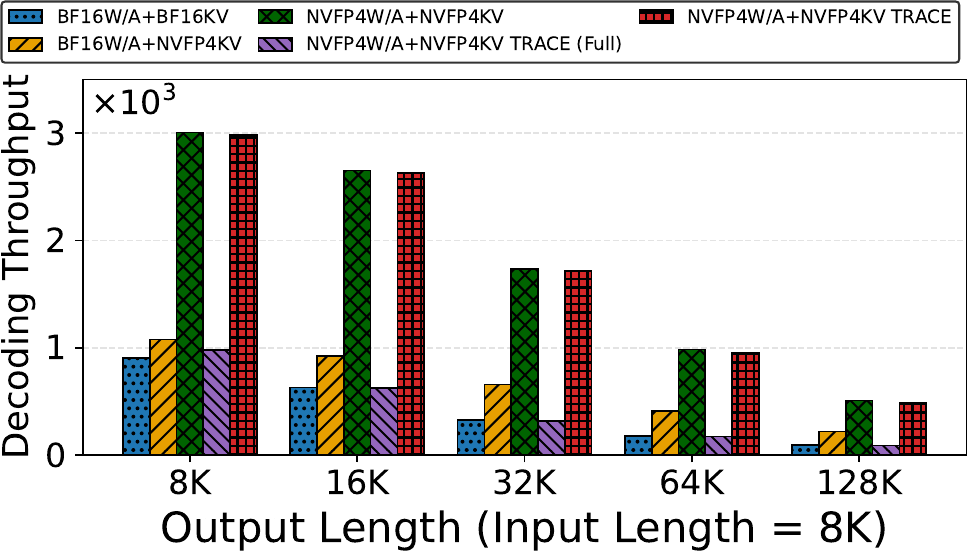}
        \label{fig:decoding_throughput}
    }
    \subfigure[E2E RL Training Step Time]{
        \centering
        \includegraphics[width=0.42\textwidth]{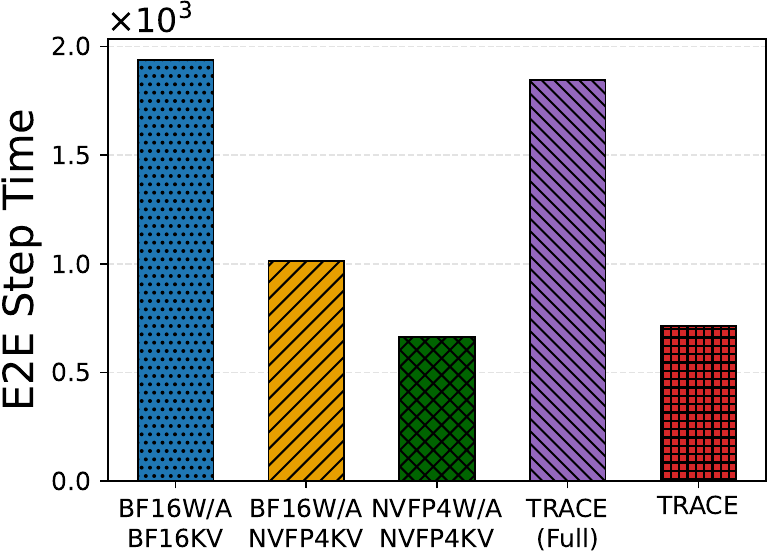}
        \label{fig:e2e_step_time}
    }
    \vspace{-2mm}
    \caption{
    Efficiency analysis of \sysname on Qwen3.5-35B-A3B.
    Decoding throughput is measured using only the rollout engine (SGLang).
    End-to-end RL training step time is measured on the reasoning RL tasks.
    }
    \label{fig:efficiency}
\end{minipage}
\end{figure}
\subsection{Efficiency Analysis}
\label{subsec:efficiency}
We next evaluate the efficiency of \sysname on Qwen3.5-35B-A3B from two aspects:
(1) rollout decoding throughput under different generation lengths, and
(2) end-to-end RL training efficiency. 
We also analyze the efficiency of Qwen3.8-Flash-Next in~\cref{appendix:flash_next_throughput}.

\begin{wrapfigure}{R}{0.26\textwidth} 
  \centering
  \vspace{-5mm}
  \includegraphics[width=\linewidth]{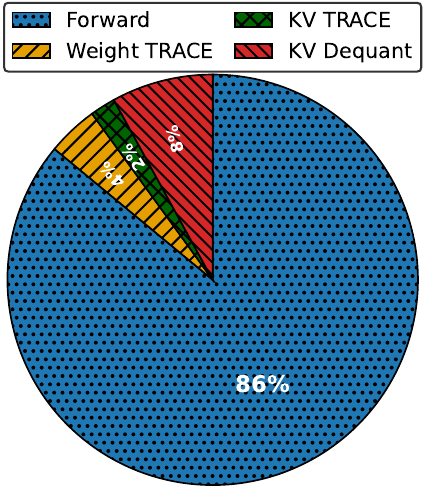} 
    \caption{Rollout time breakdown for \sysname.}
   \label{fig:breakdown}
   \vspace{-10mm}
\end{wrapfigure}
\textbf{Rollout Efficiency.}
We first evaluate the decoding throughput of \sysname under different output lengths.
As shown in~\Cref{fig:decoding_throughput}, the full rollout-guided variant incurs substantial throughput degradation due to the overhead of storing and transferring complete rollout-side quantization information.
In contrast, \sysname maintains throughput close to vanilla FP4 rollout across different output lengths.
For Qwen3.5-35B-A3B, retaining one-bit mantissa information together with amortized scale metadata for the latter 20 layers consumes approximately $2048\times20\times1.5=61{,}440$ bits, or $7.5$ KB, of cached quantization information per generated token, which reduces the cache size by approximately $6\times$ compared with caching the full quantized activation.
In particular, \sysname achieves up to $5.4\times$ higher decoding throughput at 128K output length compared with BF16 rollout, demonstrating that the proposed quantization-information caching preserves most of the decoding efficiency of FP4 rollout while enabling rollout-guided training.

\textbf{End-to-End RL Training Efficiency.}
Lastly, we evaluate the end-to-end training overhead introduced by \sysname.
As shown in \Cref{fig:e2e_step_time}, \sysname increases the RL step time from 664 to 713, corresponding to only 7.4\% overhead compared with vanilla joint FP4 rollout.
To further understand the source of this overhead, \Cref{fig:breakdown} reports the rollout-time breakdown of \sysname.
The model forward pass accounts for 86\% of the rollout time, while weight-side and KV-side rollout guidance account for only 4\% and 2\%, respectively; the remaining 8\% comes from KV dequantization.
These results demonstrate that \sysname introduces limited runtime overhead while retaining the efficiency advantage of FP4 rollout.

\WeightKVAblationTable
\subsection{Ablation Study}
\label{subsec:ablation_study}

\textbf{Isolated Weight and KV Quantization.}
We first examine the individual impact of FP4 weight and KV cache quantization.
As shown in~\cref{tab:weight_kv_ablation}, \sysname outperforms the baselines on most benchmarks in both isolated settings and consistently achieves performance comparable to BF16 rollout.
These results suggest that the joint FP4 configuration is substantially more challenging than either isolated setting, highlighting the importance of controlling the train--rollout discrepancy under joint FP4 rollout.

\PrecisionTable
\SensitivityTable
\textbf{Performance Under Microscaling FP4 Formats.}
With recent adoption of MXFP4 MegaMoE kernels for efficient MoE inference, we further examine whether \sysname generalizes to microscaling FP4 formats.
As shown in~\cref{tab:precision}, \sysname improves the average score from 69.0 to 75.1 under W4A8 MXFP4 with FP4 KV cache, and from 67.2 to 73.5 under the more aggressive W4A4 MXFP4 setting.
These results demonstrate that \sysname remains effective across different FP4 formats.

\textbf{Modular Sensitivity Study.}
Lastly, we evaluate the sensitivity of \sysname to the amount of rollout-side quantization information retained for training.
As shown in~\cref{tab:sensitivity}, retaining only one mantissa bit across all 40 layers achieves an average score of 75.8.
Keeping this information for only the latter 20 layers incurs small degradation, while reducing the coverage to 10 and 5 layers lowers the average score to 74.1 and 73.4, respectively.
These results show that \sysname preserves strong performance with only compact mantissa information from a subset of deeper layers.

\begin{figure}[h]
 \vspace{-0mm}
     \begin{minipage}{0.98\textwidth}
     \centering
    \includegraphics[width=\linewidth]{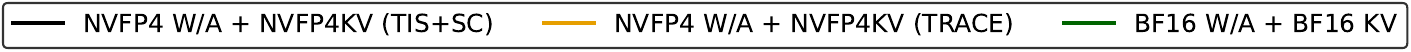}
     \end{minipage}
     \\
     \begin{minipage}{1\textwidth}
        \subfigure[Log-Prob Diff Max]{
            \centering
            \includegraphics[width=0.31\textwidth]{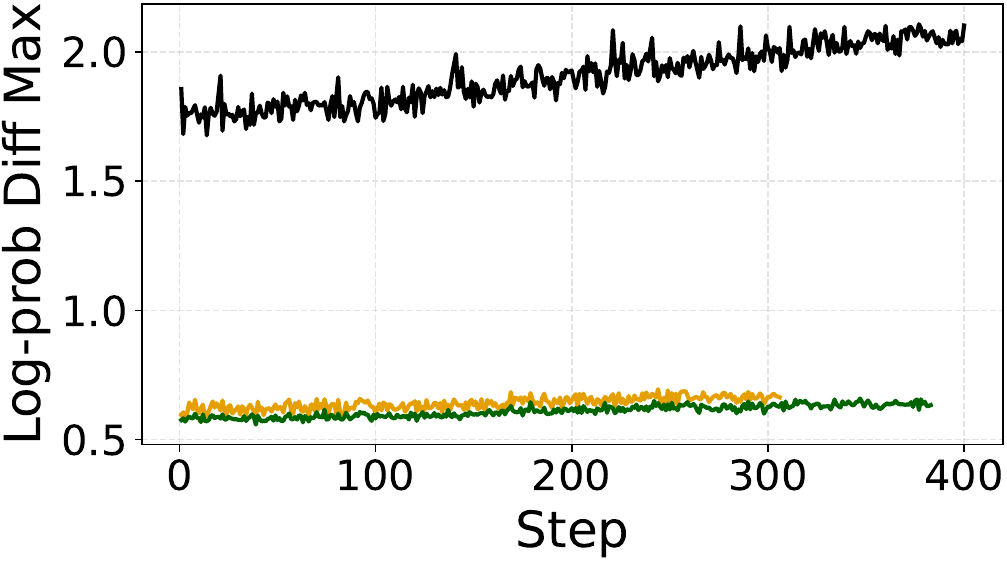}
            \label{fig:sc_logprob_diff_max}
            }
        \subfigure[Log-Prob Diff Min]{
            \centering
            \includegraphics[width=0.31\textwidth]{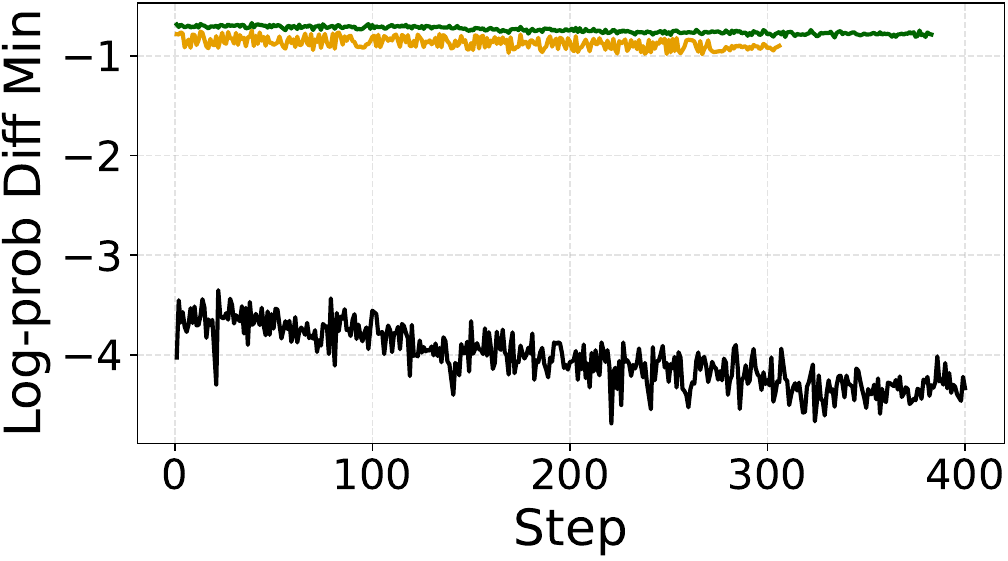}
            \label{fig:sc_logprob_diff_min}
            }
         \subfigure[Training Entropy Loss]{
            \centering
            \includegraphics[width=0.31\textwidth]{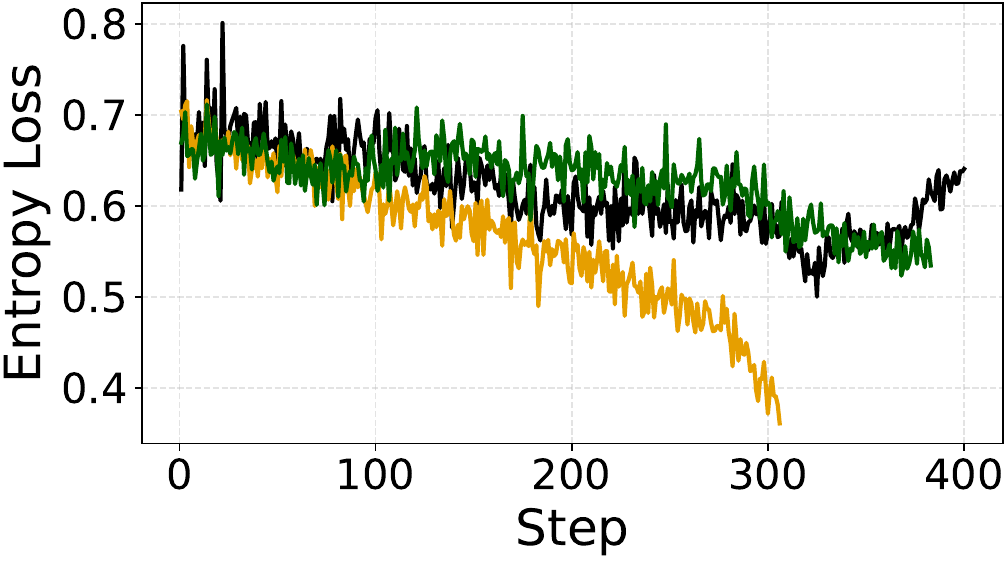}
            \label{fig:sc_training_entropy}
            }
        \\
        \subfigure[Mean Reward]{
            \centering
            \includegraphics[width=0.31\textwidth]{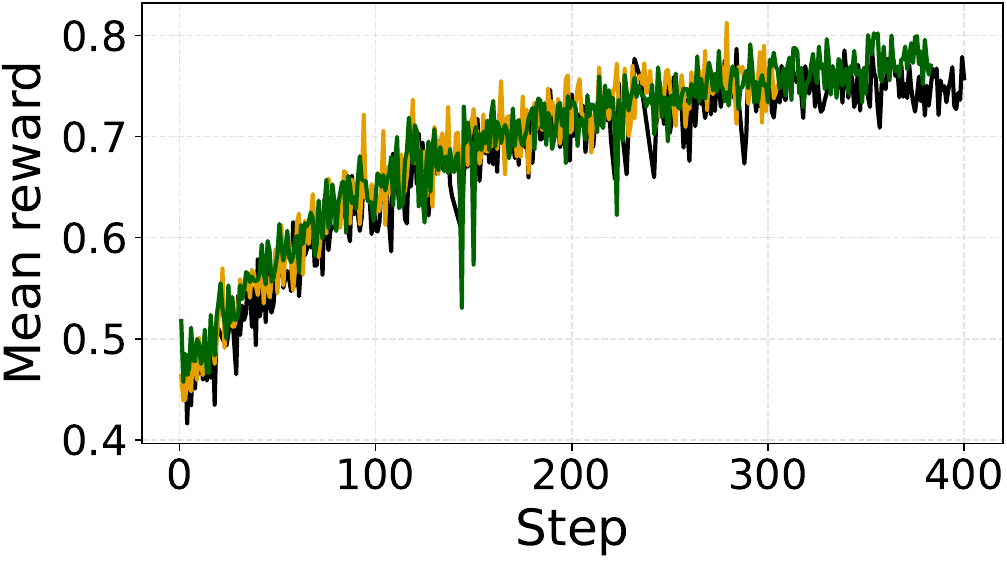}
            \label{fig:sc_reward}
            }
        \subfigure[Mean Response Length]{
            \centering
            \includegraphics[width=0.31\textwidth]{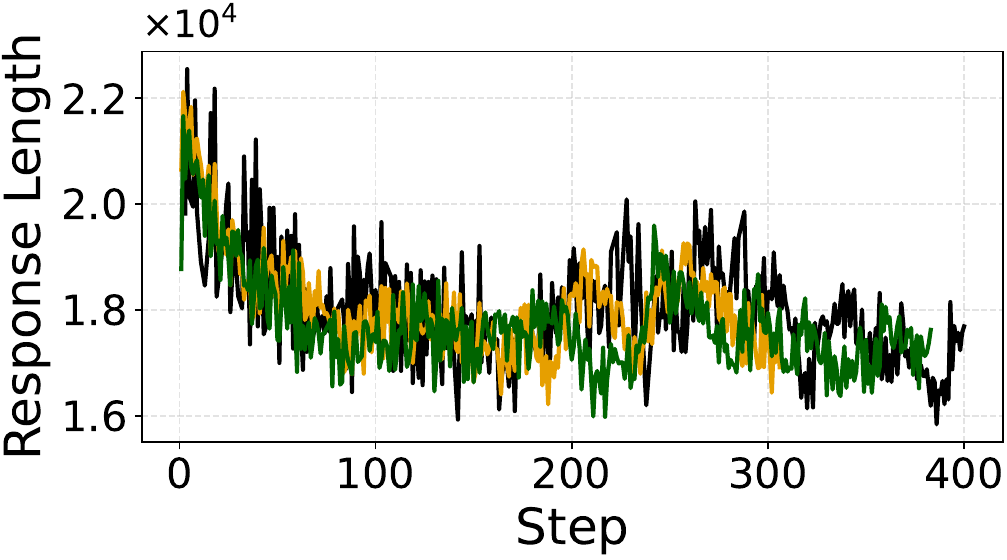}
            \label{fig:sc_response_length}
            }
        \subfigure[Test Score on HMMT]{
            \centering
            \includegraphics[width=0.31\textwidth]{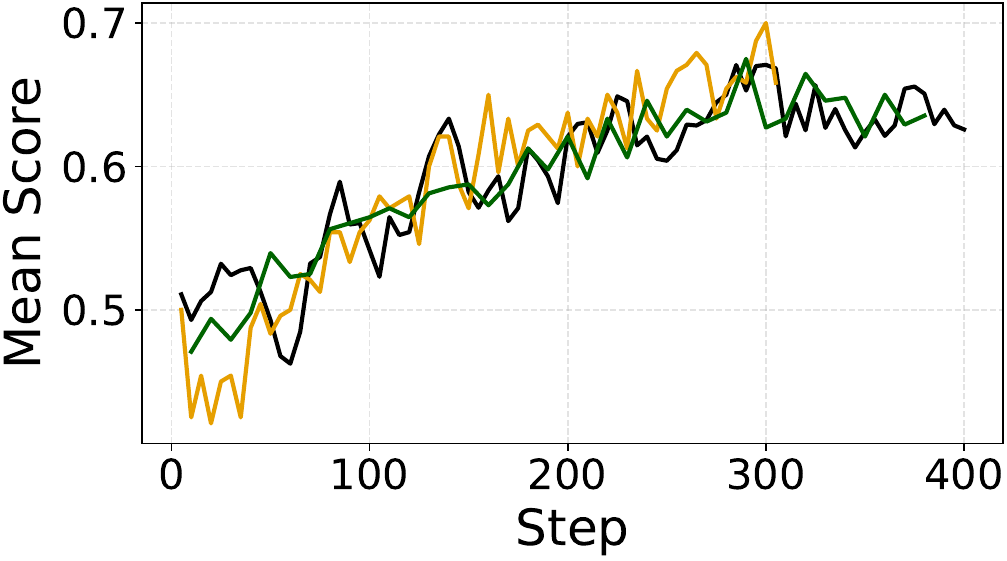}
            \label{fig:sc_val_hmmt25}
            }
        \\
        \subfigure[Test Score on AIME24]{
            \centering
            \includegraphics[width=0.31\textwidth]{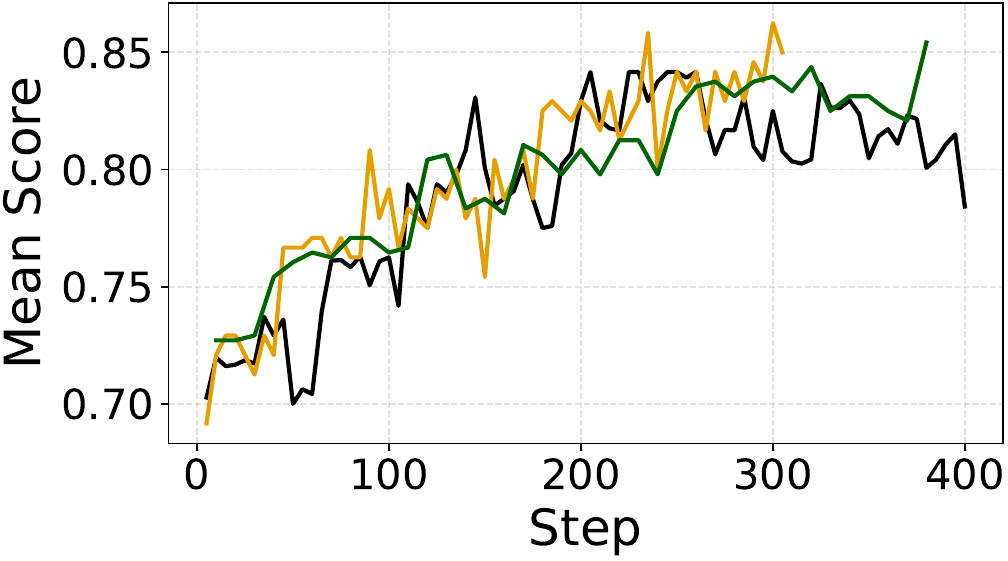}
            \label{fig:sc_val_aime24}
            }
        \subfigure[Test Score on AIME25]{
            \centering
            \includegraphics[width=0.31\textwidth]{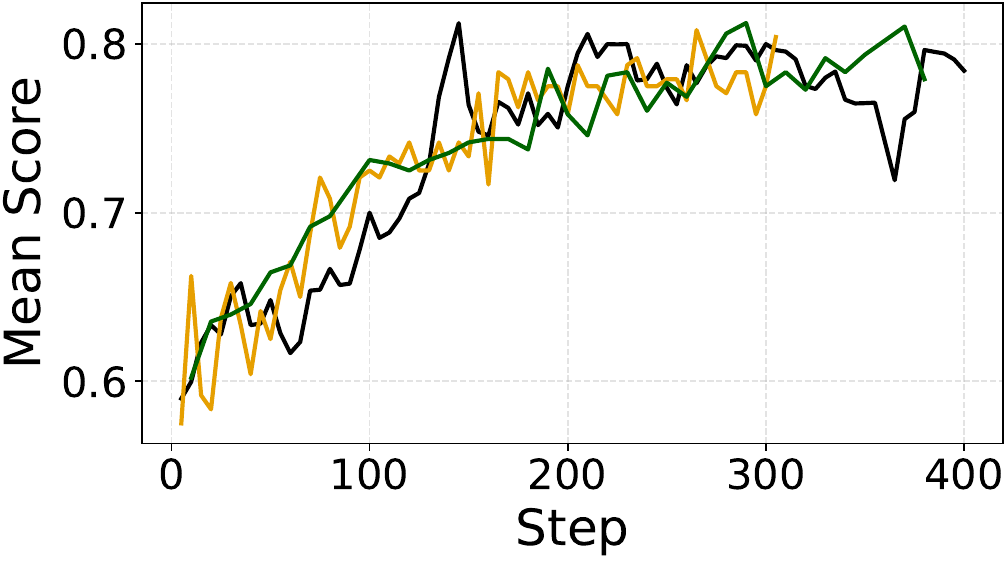}
            \label{fig:sc_val_aime25}
            }
        \subfigure[Test Score on LCB.]{
            \centering
            \includegraphics[width=0.31\textwidth]{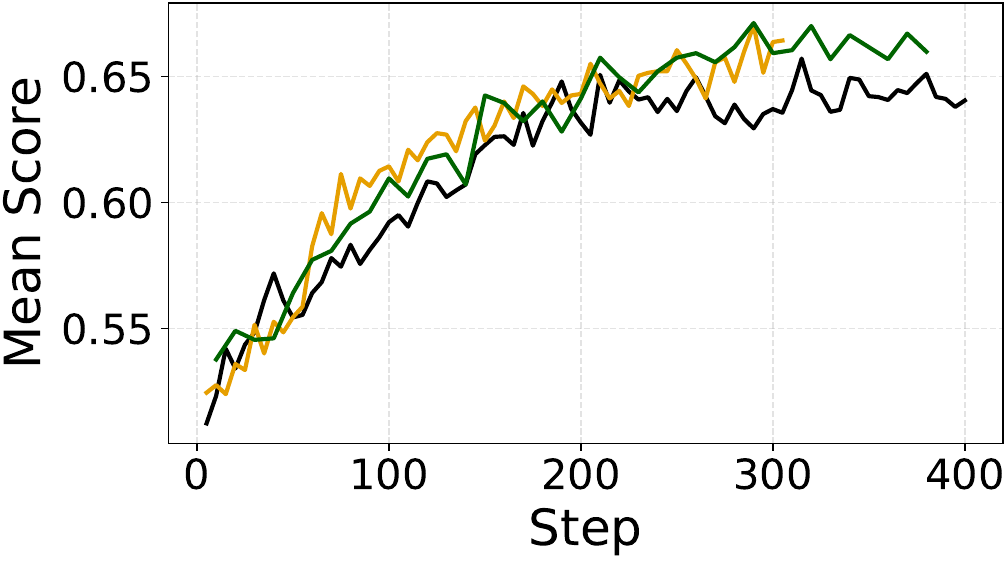}
            \label{fig:sc_val_livecodebench}
            }
        \vspace{-2mm}
        \caption{Training dynamics of Qwen3.5-35B-A3B on reasoning RL tasks, including the minimum/maximum log-probability difference, reward, entropy loss, response length, and test scores on HMMT, AIME24, AIME25, and LiveCodeBench.}
        \vspace{-2mm}
        \label{fig:score_centering}
    \end{minipage}
\end{figure}

\subsection{Comparison with Score Centering Algorithm}
\label{subsec:Comparison_with_sc}
We compare \sysname with Score Centering (SC)~\citep{marek2026score} under the same joint NVFP4 W/A/KV rollout configuration.
We replace the default policy loss with TIS+SC, combining SC with token-level truncated importance sampling (TIS) weights capped at $2$.
The centering correction is approximated using rollout policy's top-128 tokens, their probabilities under both the training and rollout policies, and the remaining probability mass.

\begin{wrapfigure}{r}{0.65\textwidth}
  \centering
  \vspace{-4mm}
  \SCTable
\end{wrapfigure}
As shown in~\cref{fig:score_centering}, Score Centering indeed prevents the severe training collapse observed with vanilla QAT as illustrated in~\cref{subsec:overall_performance}.
However, it leaves a substantially larger train--rollout log-probability discrepancy throughout training: the maximum difference remains around $1.8$--$2.1$, while the minimum stays near $-4$, both far from the BF16 reference.
In contrast, \sysname keeps the train--rollout log-probability difference close to the BF16 rollout throughout training.
We additionally compare against SC's best evaluated checkpoint, selected by the highest average score across the four benchmarks.
As shown in~\cref{tab:sc}, \sysname achieves an average score of 75.3 versus 74.1 for SC at step 380, a gain of 1.2 percentage points.
These results suggest that by directly reducing the quantization-induced discrepancy between the training and rollout paths, \sysname provides both more consistent policy alignment and stronger average task performance.

\begin{wrapfigure}{r}{0.65\textwidth} 
  \centering
  \vspace{-4mm}
  \PTQTable
\end{wrapfigure}
\subsection{Comparison with Post-Hoc FP4 Quantization}
\label{subsection:Comparison_with_ptq}
We compare the final FP4 policy performance obtained under two different training and quantization regimes.
For the PTQ baselines, RL training is first completed with BF16 rollout, after which different PTQ algorithms are applied post hoc to the resulting policy.
In contrast, \sysname performs RL training directly with joint FP4 weight/activation and FP4 KV-cache rollout, allowing the policy to adapt to low-precision execution throughout training.
As shown in~\Cref{tab:quant_weight}, the post-hoc PTQ pipelines achieve average scores of 70.4, 71.0, and 71.4 with vanilla NVFP4, 4over6, and H-Scale, respectively, whereas \sysname achieves 75.3 under FP4 rollout throughout RL training.
This result demonstrates that incorporating low-precision execution into RL training with \sysname can produce a strong final FP4 policy without relying on BF16 rollout followed by post-training quantization.

\section{Conclusion}
\label{sec:conclusion}
In this paper, we presented \sysname, an FP4 quantization framework for RL training of MoE language models.
\sysname combines rollout-guided QAT with efficient quantization-information caching to reduce train--rollout discrepancy and its associated overhead.
Experiments across multiple MoE models and RL tasks show that \sysname enables joint FP4 weight/activation and KV-cache rollout while preserving performance comparable to BF16 rollout and substantially improving rollout efficiency.



\section*{Acknowledgements}
This work was supported by Alibaba Research Intern Program.

\vspace{-2mm}


\bibliography{Reference}
\bibliographystyle{plainnat}

\appendix
\newpage
\appendix
\onecolumn
\section{Appendix.}
\subsection{Pseudocode of \sysname}
\label{appendix:pseudocode}
\cref{algo:framework} summarizes \sysname for quantized MoE inputs and KV states.
The mantissa-only setting uses $b=1$ in the latter half of the layers.
Each record contains packed low-bit rollout codes and the corresponding rollout scales, indexed by trajectory, token, and site.
During training, these records are used to construct a compact rollout-side reference for guiding the local FP4 rounding decision.

\begin{algorithm}[ht]
\caption{\sysname: Rollout-Guided FP4 QAT}
\label{algo:framework}
\begin{algorithmic}[1]
\Require Policy parameters $W$, prompts $\mathcal D$, reward function $r$, captured sites $\mathcal S$, bits $b = 1$
\Ensure Updated policy parameters $W$
\For{each RL iteration}
    \State Generate trajectories $\mathcal T$ and log-probabilities $\ell_{\mathrm{roll}}$ from $\mathcal D$ using $Q_{\mathrm{FP4}}(W)$
    \State During rollout, record $\mathcal R=\{(\operatorname{Pack}(\operatorname{low}_b(q_{\mathrm{roll}})),s_{\mathrm{roll}})\}$ at $\mathcal S$
    \State Compute advantages $A$ from rewards $r(\mathcal T)$
    \For{each policy-update minibatch $\mathcal B\subseteq\mathcal T$}
        \For{each captured activation $x$ encountered during the QAT forward}
            \State $(c,s)\gets\Call{Unpack}{\mathcal R[\text{trajectory},\text{token},\text{site}]}$
            \State $z\gets\operatorname{clip}(x/s,-6,6)$ \Comment{Normalize with the rollout scale}
            \State $\widehat q_{\mathrm{roll}}\gets\Call{Reconstruct}{c,b,z}$
            \State $(q_-,q_+)\gets\Call{BracketFP4}{z}$
            \State $q^*\gets\arg\min_{q\in\{q_-,q_+\}}|q-\widehat q_{\mathrm{roll}}|$
            \State Use $\Call{STE}{x,sq^*}$ in the remaining forward computation
        \EndFor
        \State Compute $\mathcal L_{\mathrm{RL}}$ from the QAT log-probabilities, $\ell_{\mathrm{roll}}$, and $A$ on $\mathcal B$
        \State $W\gets\Call{OptimizerStep}{W,\nabla_W\mathcal L_{\mathrm{RL}}}$
    \EndFor
\EndFor
\State \Return $W$
\end{algorithmic}
\end{algorithm}

Here $s$ combines block and token/global scales.
\textproc{Reconstruct} constructs a compact rollout-side reference by selecting the E2M1 magnitude nearest to $|z|$ whose low $b$ code bits match $c$, while using the sign of the current training activation.
Accordingly, the reconstructed value is used as a local guidance target rather than as an exact recovery of the original rollout-side FP4 codeword.
\textproc{BracketFP4} returns the adjacent E2M1 rounding candidates around $z$, and \textproc{STE} returns its second argument in the forward pass while passing gradients to its first argument.

\subsection{Training Dynamics on Qwen3.8-Flash-Next}
\label{appendix:coding_training_curves}
\begin{figure}[ht]
    \centering
    \includegraphics[width=0.98\linewidth]{Figures/fp4_kv/legend.pdf}\\
    \subfigure[Log-Prob Diff Max]{
        \includegraphics[width=0.31\textwidth]{Figures/fp4_kv/actor_mixed_logprob_diff_max.pdf}
        \label{fig:coding_logprob_diff_max}
    }
    \subfigure[Log-Prob Diff Min]{
        \includegraphics[width=0.31\textwidth]{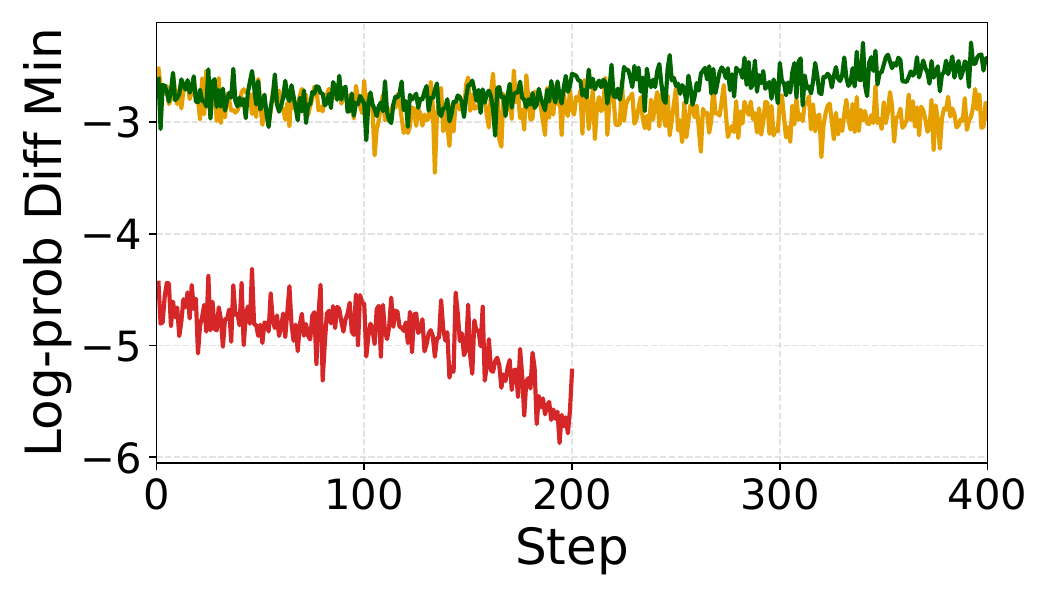}
        \label{fig:coding_logprob_diff_min}
    }
    \subfigure[Mean Reward]{
        \includegraphics[width=0.31\textwidth]{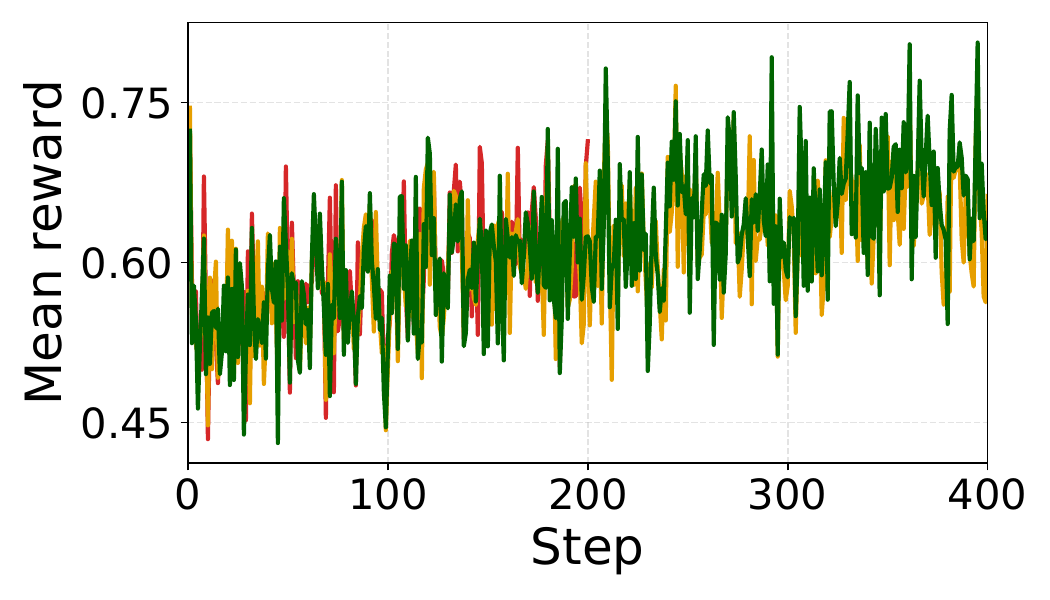}
        \label{fig:coding_reward}
    }\\
    \subfigure[Training Entropy Loss]{
        \includegraphics[width=0.31\textwidth]{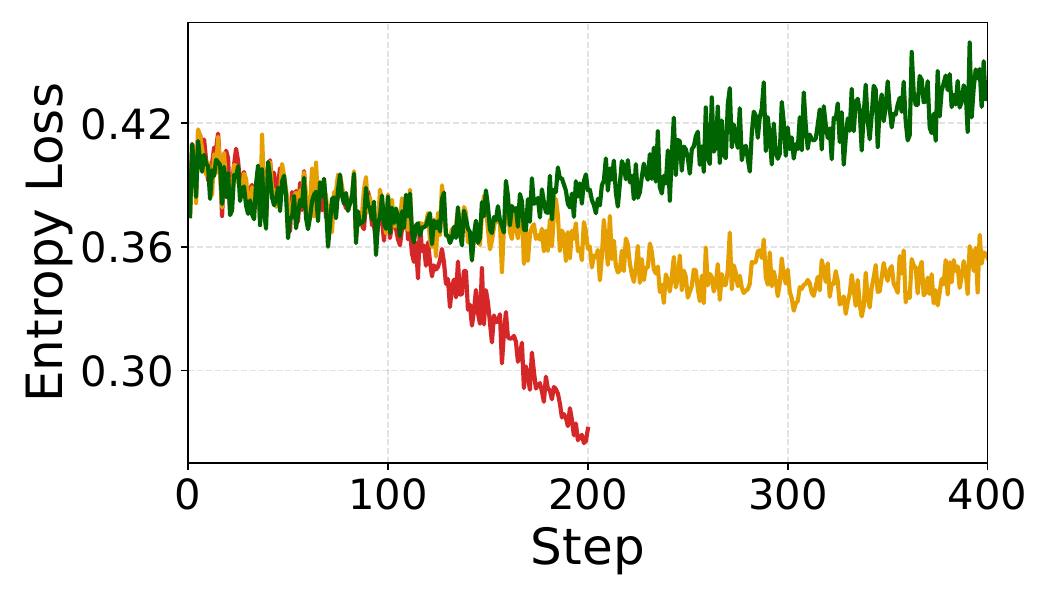}
        \label{fig:coding_training_entropy}
    }
    \subfigure[Mean Response Length]{
        \includegraphics[width=0.31\textwidth]{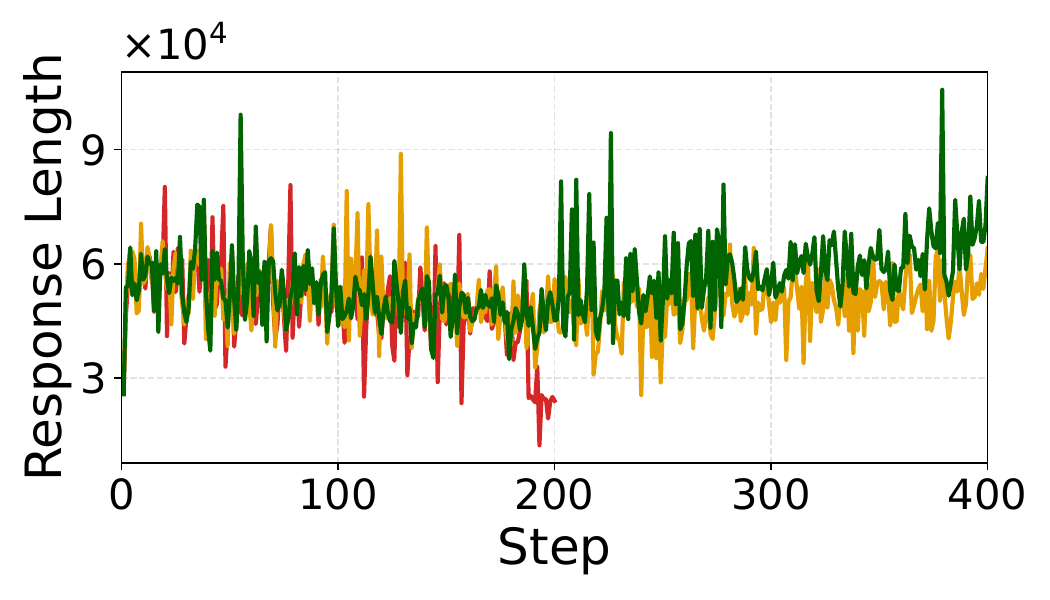}
        \label{fig:coding_response_length}
    }
    \subfigure[Test Score on TB]{
        \includegraphics[width=0.31\textwidth]{Figures/fp4_kv/val_tb_mean.pdf}
        \label{fig:coding_val_tb}
    }
    \vspace{-2mm}
    \caption{Performance evolution of Qwen3.8-Flash-Next on coding RL tasks: maximum and minimum log-probability differences, mean reward, training entropy loss, mean response length, and test score on Terminal-Bench~2.1.}
    \label{fig:coding_training_curves}
    \vspace{-2mm}
\end{figure}

\Cref{fig:coding_training_curves} complements the main-text analysis of coding RL on Qwen3.8-Flash-Next. We compare vanilla QAT and \sysname under joint NVFP4 weight and KV-cache quantization with a BF16 rollout reference. The six panels show the same training diagnostics as in the main text, together with test scores on Terminal-Bench~2.1 (TB), over a window of 400 training steps.

\textbf{Training--Rollout Consistency.}
In the recorded training steps, vanilla QAT exhibits larger log-probability differences than both \sysname and the BF16 reference. Its maximum difference increases and its minimum becomes more negative toward the end of the recorded run. In contrast, \sysname remains much closer to the BF16 reference in both diagnostics, with no comparable widening over the 400-step window. This behavior is consistent with the improved training--rollout alignment observed on reasoning RL tasks.

\textbf{Reward, Entropy, and Response Length.}
Vanilla QAT nevertheless exhibits a pronounced entropy decrease after approximately step 100 and shorter responses near the end of its recorded trajectory. \sysname shows a more gradual entropy decrease, maintains long responses, and improves reward on average. Its recorded TB test scores also increase over the training window. These observations support the training-stability trend in the main text.

\subsection{Training Dynamics on Qwen3.8-2.4T-A95B}
\label{appendix:max_training_curves}
\begin{figure}[htbp]
    \centering
    \includegraphics[width=0.98\linewidth]{Figures/fp4_kv/legend.pdf}\\
    \subfigure[Log-Prob Diff Max]{%
        \includegraphics[width=0.235\textwidth]{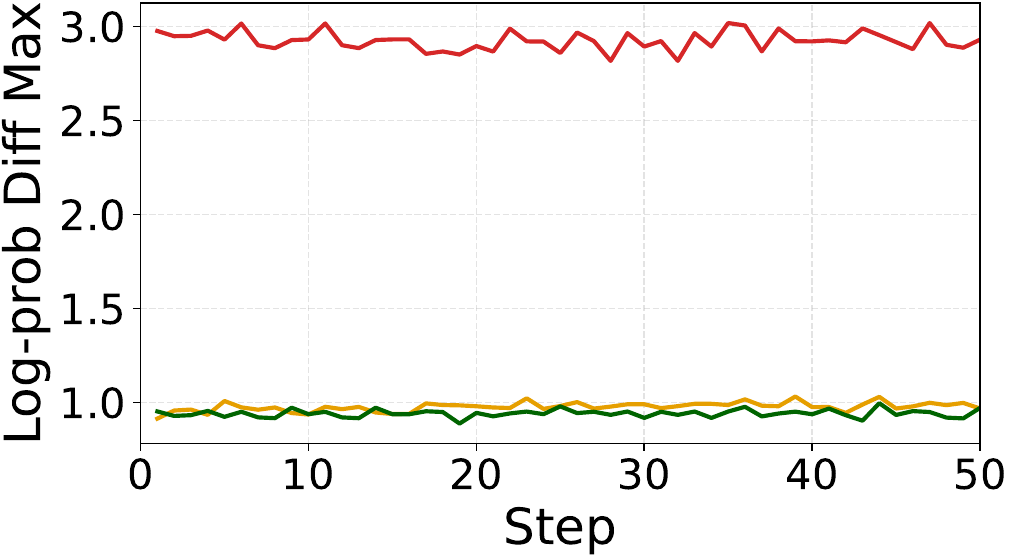}%
        \label{fig:max_logprob_diff_max}%
    }\hfill%
    \subfigure[Mean Reward]{%
        \includegraphics[width=0.235\textwidth]{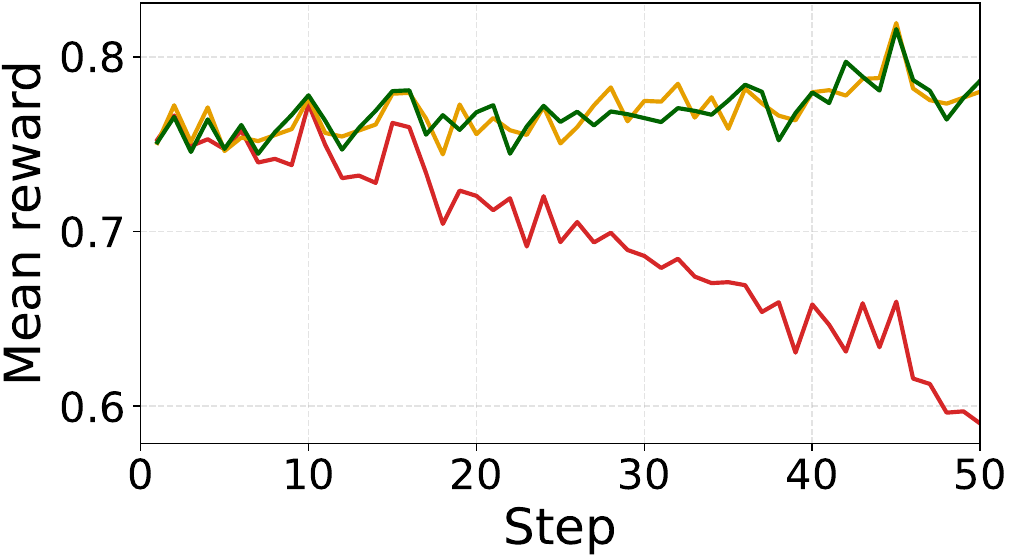}%
        \label{fig:max_reward}%
    }\hfill%
    \subfigure[Entropy Loss]{%
        \includegraphics[width=0.235\textwidth]{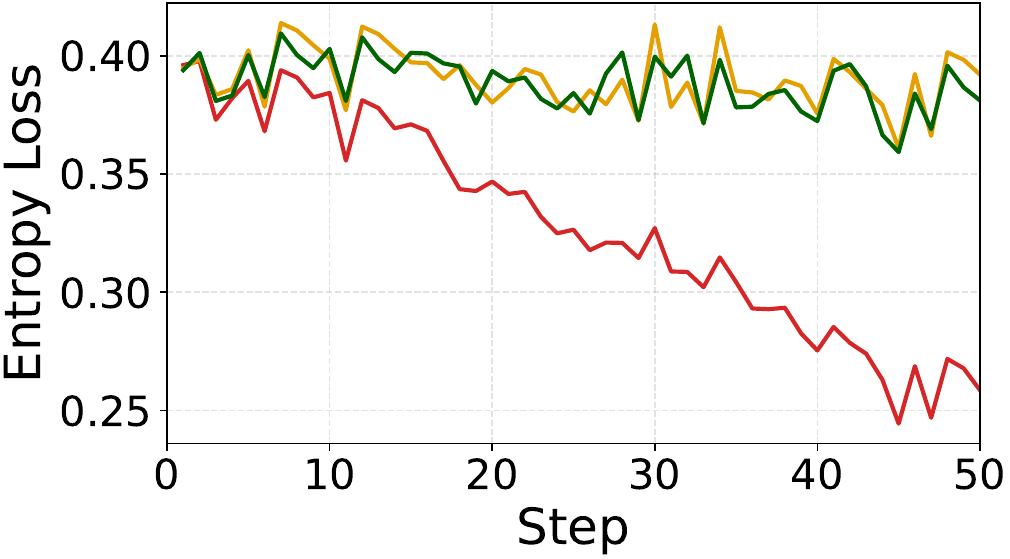}%
        \label{fig:max_training_entropy}%
    }\hfill%
    \subfigure[Response Length]{%
        \includegraphics[width=0.235\textwidth]{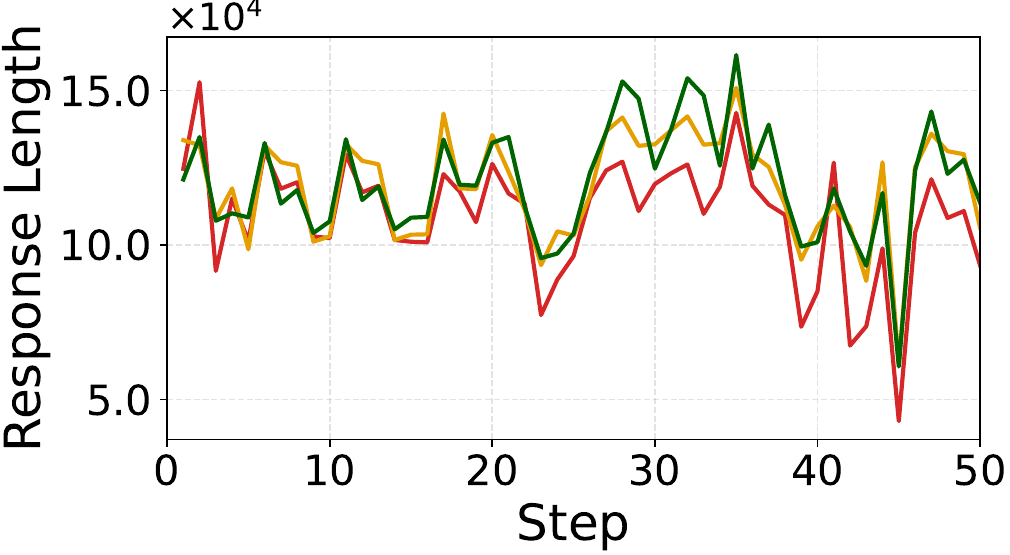}%
        \label{fig:max_response_length}%
    }
    \vspace{-2mm}
    \caption{Training dynamics for Qwen3.8-2.4T-A95B (Max): (a) maximum log-probability difference, (b) mean reward, (c) training entropy loss, and (d) mean response length.}
    \label{fig:max_training_curves}
    \vspace{-2mm}
\end{figure}

\Cref{fig:max_training_curves} complements the main-text comparison of long-horizon RL on Qwen3.8-2.4T-A95B. We compare vanilla QAT and \sysname under joint NVFP4 weight and KV-cache quantization with a BF16 rollout reference. The four panels report the maximum log-probability difference, mean reward, training entropy loss, and mean response length over the 50-step training run.

\begin{wrapfigure}{R}{0.52\textwidth} 
  \centering
    \includegraphics[width=\linewidth]{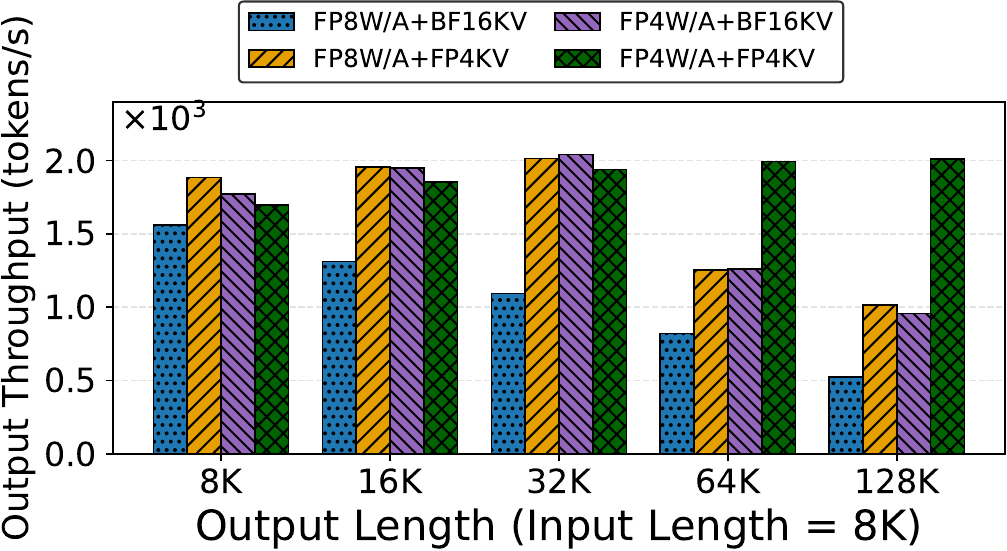}
    \caption{Qwen3.8-Flash-Next output throughput under four expert-weight/KV precision combinations. All settings use one GPU, BF16 PLE offload, 8K-token inputs, and 64 submitted requests.}
    \label{fig:flash_next_throughput}
   \vspace{-5mm}
\end{wrapfigure}

\textbf{Training--Rollout Consistency.}
Over the 50 recorded steps, vanilla QAT keeps a larger maximum log-probability difference, near 2.9, while \sysname and the BF16 reference stay near 0.95. \sysname remains close to the BF16 reference throughout the window. This behavior is consistent with the improved training--rollout alignment observed on reasoning and coding RL tasks.

\textbf{Reward, Entropy, and Response Length.}
For Vanilla QAT, the mean reward falls from about 0.75 to about 0.59, training entropy falls from about 0.40 to about 0.26, and mean response length shortens from about 125K to about 93K tokens. \sysname raises mean reward from about 0.75 to about 0.78, keeps entropy near the BF16 reference at about 0.38--0.40, and keeps responses in the 106K--134K range. These observations support the training-stability trend reported for long-horizon RL in the main paper.

\subsection{Throughput Analysis on Qwen3.8-Flash-Next}
\label{appendix:flash_next_throughput}

\textbf{Setup and Metric.}
Following~\cref{fig:decoding_throughput}, we compare four routed-expert/KV precision combinations on Qwen3.8-Flash-Next: FP8/BF16, FP8/FP4, FP4/BF16, and FP4/FP4. All use the same checkpoint on one GPU, BF16 PLE/ngram embeddings offloaded to CPU pinned memory, and no multi-token prediction. The GPU memory fraction is 0.85, with automatic KV allocation. Each measurement submits 64 synthetic 8K-token prompts and generates exactly 8K--128K output tokens per request. 

\textbf{Dependence on Output Length.}
In~\cref{fig:flash_next_throughput}, joint FP4 improves throughput over FP8/BF16 by $1.09\times$, $1.41\times$, $1.77\times$, and $2.43\times$ at 8K, 16K, 32K, and 64K outputs, respectively. In the single-run 128K supplement, joint FP4 reaches 2013 tokens/s versus 526 tokens/s for FP8/BF16 ($3.83\times$). The corresponding peak decoding batches are 64 and 7 requests, so this gain includes the concurrency enabled by KV capacity.

\end{document}